\documentclass[11pt]{article}

\usepackage[margin=1in]{geometry}
\usepackage[T1]{fontenc}
\usepackage[utf8]{inputenc}

\usepackage{graphicx}
\usepackage{comment}
\usepackage{mathtools}
\usepackage{multirow}
\usepackage{url}
\usepackage{soul}
\usepackage{amsmath,amssymb,amsfonts}
\usepackage{booktabs}
\usepackage{textcomp}
\usepackage{xcolor}
\usepackage{float}
\usepackage{subfig}
\usepackage[final]{microtype}
\usepackage[colorlinks=true,linkcolor=blue,citecolor=blue,urlcolor=blue]{hyperref}

\title{Efficient Semantic Image Communication for Traffic Monitoring at the Edge}

\author{
Damir Assylbek$^{1}$, Nurmukhammed Aitymbetov$^{1}$, Marko Ristin$^{2}$, and Dimitrios Zorbas$^{1}$\\[0.5em]
\small $^1$ Nazarbayev University, School of Engineering \& Digital Sciences, Astana, Kazakhstan \\
\small $^2$ Zurich University of Applied Sciences (ZHAW), Winterthur, Switzerland
}

\date{}

\begin{document}

\maketitle

\begin{abstract}
Many visual monitoring systems operate under strict communication constraints, where transmitting full-resolution images is impractical and often unnecessary. In such scenarios, visual content is typically analyzed for object presence, spatial relationships, and scene context rather than exact pixel accuracy. This paper presents two semantic image communication pipelines, MMSD and SAMR, that reduce transmission cost while preserving meaningful visual information for traffic monitoring. MMSD (Multi-Modal Semantic Decomposition) is designed for high compression together with data confidentiality, as sensitive data is not transferred. It replaces the original image with compact semantic representations, namely segmentation maps, edge maps, and textual descriptions, and reconstructs the scene at the receiver using a diffusion-based generative model. SAMR (Semantic-Aware Masking Reconstruction) is designed for higher visual quality preservation without sacrificing compression rate. It selectively suppresses non-critical image regions according to semantic importance before standard JPEG encoding, and restores the missing content at the receiver through generative inpainting. Both designs follow an asymmetric sender-receiver architecture, where lightweight processing is performed at the edge and computationally intensive reconstruction is offloaded to the server. Measured on a Raspberry Pi~5, the edge-side processing time is approximately 15\,s for MMSD and 9\,s for SAMR. Moreover, SAMR complements MMSD by enabling on-demand, higher-fidelity transmission of semantically important regions with minimal additional edge-side overhead, since it reuses the same segmentation output and computes masking very efficiently. Experimental evaluation showed 99\% and 99.1\% average reductions in transmitted data for MMSD and SAMR, respectively. Furthermore, MMSD achieved lower payload size than the recent SPIC baseline while preserving strong semantic consistency, whereas SAMR provided a better quality-compression trade-off than standard JPEG and the SQ-GAN baseline under comparable operating conditions.
\end{abstract}


\section{Introduction}

The widespread adoption of visual sensing in bandwidth-constrained environments--such as smart cities \cite{el2024artificial}, traffic surveillance systems \cite{aminiyeganeh2024iot}, unmanned aerial platforms \cite{bakirci2025internet}, and Internet of Things (IoT) deployments \cite{kargar2025tiny}--has intensified the demand for efficient image transmission mechanisms. In these settings, communication links are often the primary bottleneck, while the transmitted visual data is primarily consumed by automated systems or human operators interested in high-level scene understanding rather than pixel-perfect fidelity.

Traditional image compression methods, such as JPEG and Motion-JPEG, remain widely used due to their simplicity and low computational requirements, but they suffer from severe blocking and blurring artifacts at low bitrates, often degrading semantically important content \cite{survey2025_vaps}. More recently, learned compression techniques based on variational autoencoders and generative models have improved rate--distortion performance by optimizing pixel-level fidelity; however, they remain fundamentally syntactic and struggle to preserve object-level semantics at very low bitrates \cite{chen2024generative,giuliano2024convolutional,lu2024hybridflow}. To address this limitation, semantic communication has emerged as a promising paradigm that focuses on transmitting task-relevant information, leading to semantic-aware compression approaches that leverage region-of-interest masks or segmentation \cite{luo2022semantic,jiang2024tlic,wang_semantic-aware_2024,shanmugam2025semantic}. In parallel, generative reconstruction methods exploit semantic priors, such as segmentation maps, to recover missing content at the receiver \cite{chang_semantic-aware_2023,gu2024semantic}. Despite these advances, existing approaches often increase sender-side complexity, rely on specialized codecs, or assume symmetric computational capabilities, limiting their practicality in resource-constrained edge environments.

This paper proposes two complementary approaches for traffic monitoring, each targeting a different operational objective. The first contribution is a multi-modal semantic decomposition (MMSD) framework for very high compression, in which the image is replaced by compact semantic representations (segmentation maps, textual scene descriptions, and edge maps), and then reconstructed at the receiver. In particular, the use of edge maps provides explicit structural guidance that has not been explored in previous semantic image communication frameworks in the literature, helping preserve object boundaries and scene layout under extreme compression. The second contribution is a semantic-aware masking reconstruction (SAMR) framework for higher visual quality preservation, which exploits semantic importance information to selectively remove non-critical image patches before standard JPEG encoding. This reduces bitrate while retaining semantically important content with higher fidelity, and the missing regions are subsequently restored at the receiver through generative inpainting. To the best of our knowledge, semantic-aware masking has not been explored in the literature.

These two designs enable flexible deployment across traffic monitoring scenarios. For example, the high-compression approach is suitable for applications where bandwidth is severely constrained and only high-level scene understanding is required, such as congestion estimation \cite{yu2024realtime}, traffic density monitoring \cite{jian2025selective}, incident detection \cite{pomoni2025smart}, or privacy-sensitive city-wide deployments \cite{wang2024privacy}. In contrast, the higher-quality approach is more appropriate for applications that require clearer visual detail in regions of interest, such as traffic violation analysis \cite{hernandezdiaz2025motorcycle}, operator-assisted surveillance \cite{stiller2026webcam}, and smart intersections \cite{mohamed2025multicamera}, where vehicles and pedestrians must be preserved more faithfully.

In summary, the key contributions of this paper are as follows:
\begin{enumerate}
    \item We design MMSD, a multi-modal semantic decomposition approach that transmits only compact semantic representations, combining segmentation maps, textual scene descriptions, and edge maps, the latter providing explicit structural guidance that has been largely unexplored in prior semantic image communication frameworks.
    \item We introduce SAMR, a semantic-aware masking framework that leverages the semantic importance information to selectively remove non-critical image patches prior to standard image encoding, enabling substantial bitrate reduction while keeping sufficient information to reconstruct the missing content back at the receiver side via generative inpainting.
    \item We conduct an extensive evaluation on the publicly available datasets, combining pixel-level metrics, compression efficiency analysis, human perceptual studies, and large vision-language model (VLM)–based semantic assessment, demonstrating significant reductions in transmitted data while maintaining object-level semantic consistency.
\end{enumerate}

The remainder of this paper is organized as follows: Section~\ref{sec:rw} reviews related work on image compression and generative reconstruction. Section~\ref{sec:multimodal} presents the proposed multi-modal semantic decomposition framework for high compression. Section~\ref{sec:masking} introduces the semantic-aware masking framework designed for higher visual quality preservation. Section~\ref{sec:eval} describes the experimental setup and discusses the evaluation results of the two approaches. Finally, Section~\ref{sec:conclusion} concludes the paper and outlines directions for future work.

\section{Related Work}
\label{sec:rw}

This section reviews recent approaches and provides the current status of research for efficient image compression, understanding and reconstruction techniques, highlighting their relevance to the proposed solution. 

\subsection{Image Compression}

The concept of image compression recognizes that different image regions contribute differently to perceptual quality and downstream task performance. 
Traditional image compression is based on signal processing, leveraging frequency transforms and analysis, where diffuse regular areas containing few or now edges (low frequency) are compressed away, while the regions with many edges (high frequency) are retained.
Since there is no training or modeling involved, implementing and deploying compression based on frequencies is straightforward.
Given its simplicity, it is the base for the currently popular image compressions such as JPEG and PNG~\cite{survey2025_vaps}.

More recent works try to learn the compression model instead of relying on the frequency analysis.
Current state-of-the-art leverages neural networks, where the network is tasked to learn to map an image to a latent code (as a compressed, shorter representation), and then reproduce the original image from the code~\cite{generativeCompression}.
This approach is extended in many directions: by adding a jitter to the code so that the network is nudged to map visually similar images to proximate codes (variational auto-encoders, VAEs); training a discriminator network to penalize visually unexpected re-constructions (generative adversarial networks, GANs); or diffusion models, where the image is mapped to a code, and then reconstructed through learned reverse diffusion steps.
For a thorough review, we point to~\cite{generativeVisualCompression}.

Usually, the networks learn to operate purely on appearance.
However, semantics, the \emph{meaning}, of the image is often crucial for the compression, where regions of interest (RoIs) with application-relevant meaning must be better preserved than irrelevant regions.
To incorporate the semantics, Gu~et~al.~\cite{rudinac_semantic_2024} propose to condition the networks with importance maps, ensuring that image regions with similar semantic importance receive similar treatment.
Similarly, Chang~et~al.~\cite{chang_semantic-aware_2023} introduce a semantic-aware visual decomposition based on semantic maps and textures, where different compression rates are applied for relevant and irrelevant regions.
Wang~et~al.~\cite{wang_semantic-aware_2024} developed a semantic-aware architecture that independently compresses relevant and non-relevant segments with different compression qualities.
In the context of autonomous driving, Shanmugam and Maheswari~\cite{shanmugam2025semantic} applied pixel-level semantic segmentation to video compression, employing different quantization factors for relevant and irrelevant regions.

\begin{figure*}
    \centering
    \includegraphics[width=522pt]{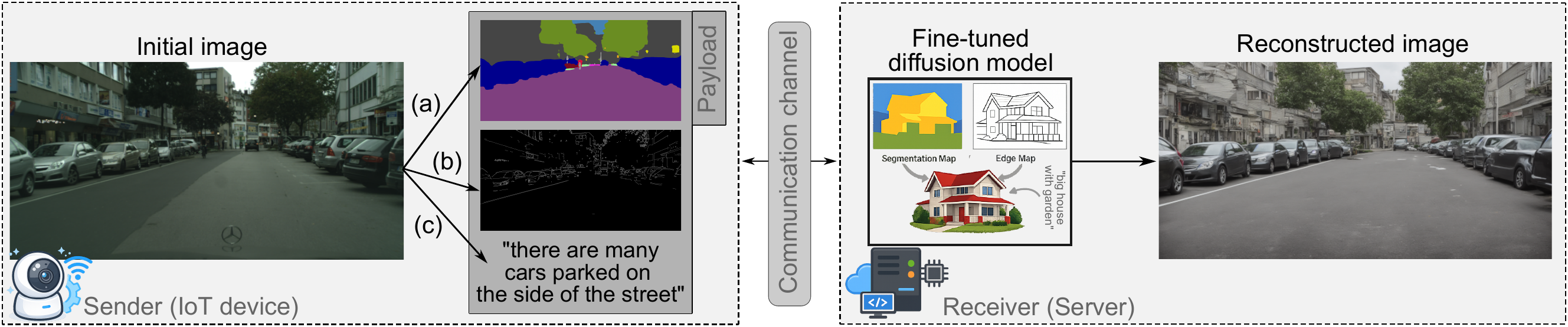}
    \caption{Pipeline with semantic decomposition and generative reconstruction. The payload consists of (a) the segmentation map, (b) the edge map, and (c) a short description (caption) of the initial image.}
    \label{fig:pipeline2}
\end{figure*}

Important to these methods is semantic segmentation, i.e., dividing an image into meaningful regions and assigning each region a label (such as road, building, or pedestrian) to understand its role and importance in the scene.
A key challenge in this task is capturing both fine-grained local details and long-range contextual relationships across distant parts of an image.
Recently, neural network approaches based on transformer architectures have shown strong performance on this problem~\cite{semanticSegmentationSurvey}.
They work by splitting an image into small patches and allowing the model to compare every patch with every other patch, so that information from the entire image can be considered when deciding the label of each region.
The training is performed in a supervised manner, where a class label is predicted for every pixel in an image.

Our approach combines semantic segmentation based on transformers and RoI-aware compression in a lightweight pipeline.
In contrast to works operating on whole image, we decouple semantic prioritization from the compression mechanism itself.
Instead of integrating semantic awareness into learned codecs or distortion objectives, we selectively mask non-critical regions prior to standard encoding.
Thus we preserve compatibility with existing infrastructures and significantly reduce sender-side complexity.

\subsection{Image Reconstruction}

Given a compressed image, image reconstruction tries to re-produce its original form from it.
As a matter of fact, this is the twin task of the (semantic and visual) compression, as the neural networks such as auto-encoders or diffusion models can be precisely trained for the reconstruction of incomplete, compressed images~\cite{generativeVisualCompression}.
For example, random masks of training images are occluded during training, and the network learns to reconstruct the occlusions~\cite{maskedAutoencoders}.

This process can be further aided by supplying additional semantic information.
For instance, Semantic-Preserving Image Coding (SPIC) by Pezone~\cite{pezone2025semantic} combines semantic segmentation maps with low-resolution coarse image files to guide diffusion-based decoders.
Similarly, SQ-GAN~\cite{Pezone2025sqgan} obtains first the semantic segmentation as well as a codebook of patterns.
Based on the application-specific importance of different regions, the method then selectively transmits the quantized patches according to the codebook.
Alternatively, other visual cues can be used to guide the reconstruction, such as object edges~\cite{softEdgeSuperResolution}. 

Textual semantics have also been explored as an additional semantic channel.
Frameworks such as MISC~\cite{wu2025semantics} incorporate large multimodal models (e.g., GPT-4V \cite{openai_gpt4_research}) to extract descriptive captions that summarize object categories and scene context. 
It provides complementary high-level guidance for the generative decoder. 
However, text-based guidance alone often suffers from spatial ambiguity, since natural language cannot precisely encode object boundaries or layout.
For example,  Xue~et~al.~\cite{xue2025onestepdiffusionbasedimagecompression} argue that purely textual prompts may be insufficient to capture fine-grained visual semantics, and instead proposes spatially aligned semantic cues (derived from an auto-encoder) to support one-step diffusion reconstruction.
In particular, this observation motivates the need for additional geometric constraints beyond language alone.

We combine the existing approaches to a novel trimodal method, where text, semantic and edge map are combined as guidance for reconstruction model.
These maps are important information-frugal contextual cues which can be highly compressed themselves due to their large connected regions and limited palette.

\section{MMSD: Multi-Modal Semantic Decomposition with Edge Maps and Captions}
\label{sec:multimodal}

This section introduces MMSD, which operates entirely in the semantic domain by decomposing each image into compact structural and contextual representations (segments, edges, and caption) that are transmitted instead of pixel data, drastically reducing the data transfer size, while preserving data confidentiality. At the receiver, these semantic modalities are used to regenerate a visually coherent image through diffusion-based reconstruction, maximizing compression by sacrificing pixel-level fidelity in favor of preserving object-of-interest structure and scene layout. 


Unlike most prior approaches that rely mainly on segmentation or low resolution original images, MMSD also incorporates edge maps as an explicit structural cue. As mentioned earlier, combined with segmentation and captions, edge maps help preserve object boundaries, scene layout, and high-level context more effectively than heavily compressed pixel data, which often suffers from blur and artifacts at low bitrates.

\subsection{Overall Pipeline}

\begin{figure*}
\centering
\includegraphics[width=250pt]{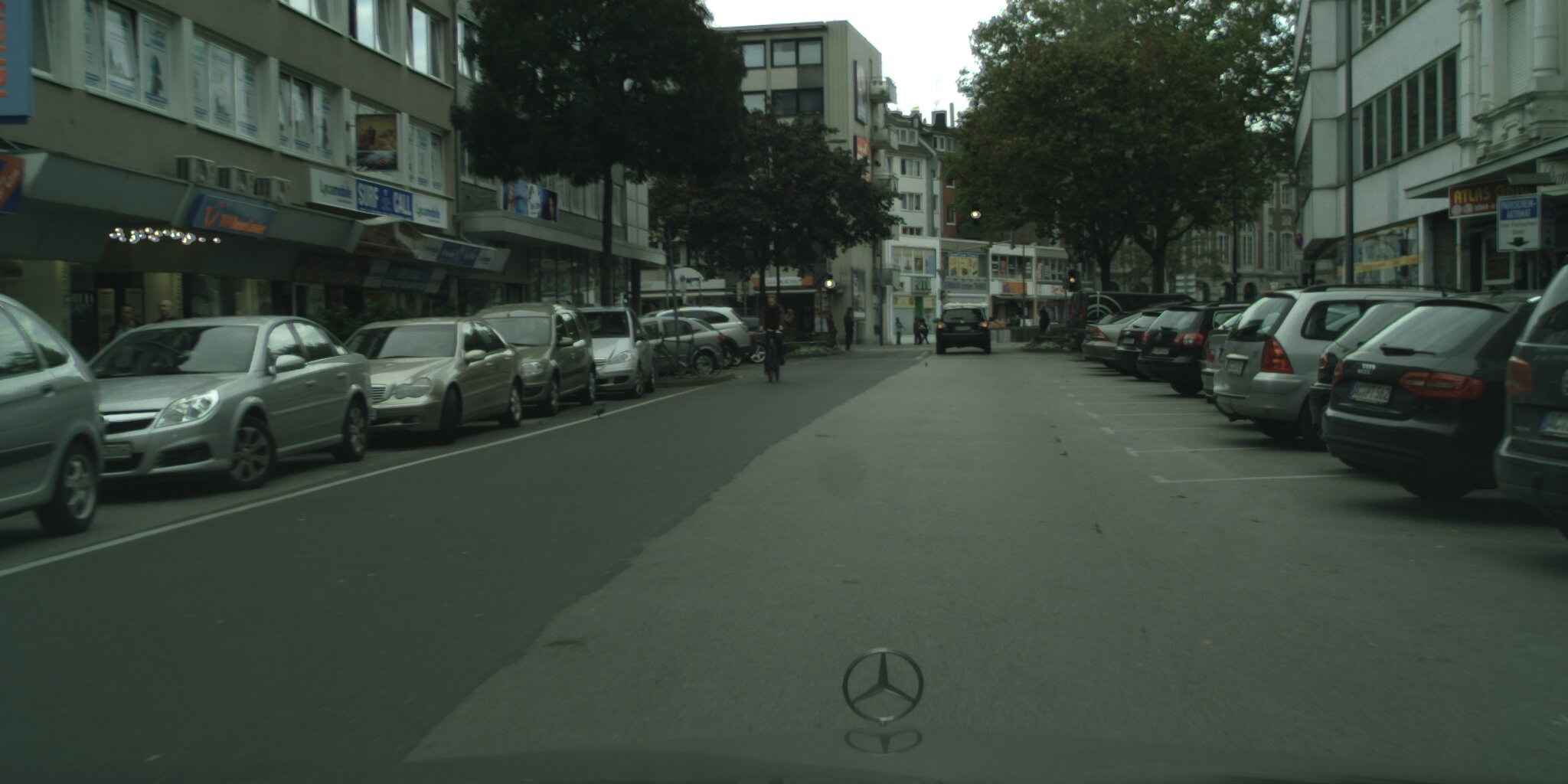}\\
\vspace{2pt}
\includegraphics[width=220pt]{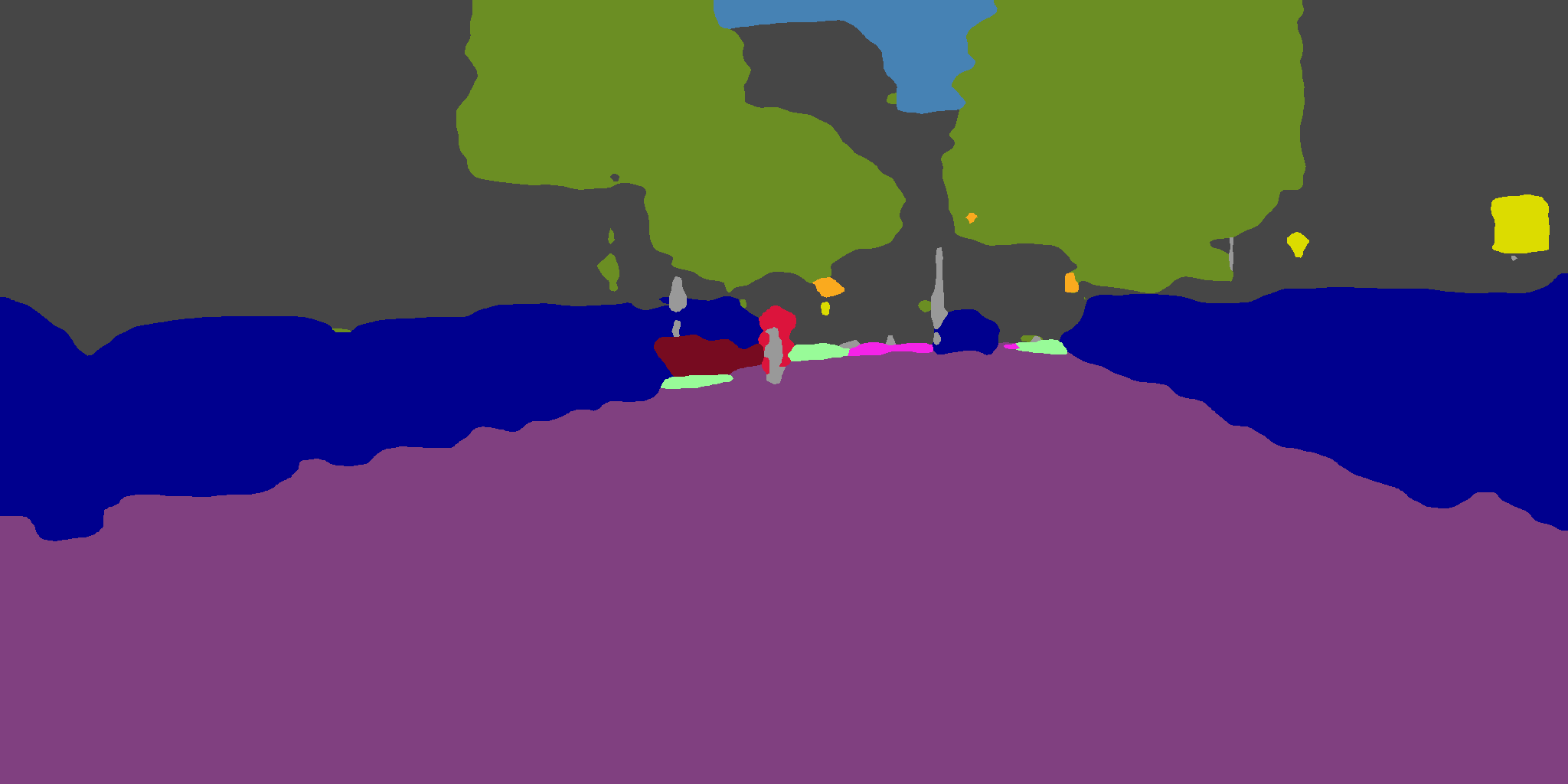}
\includegraphics[width=220pt]{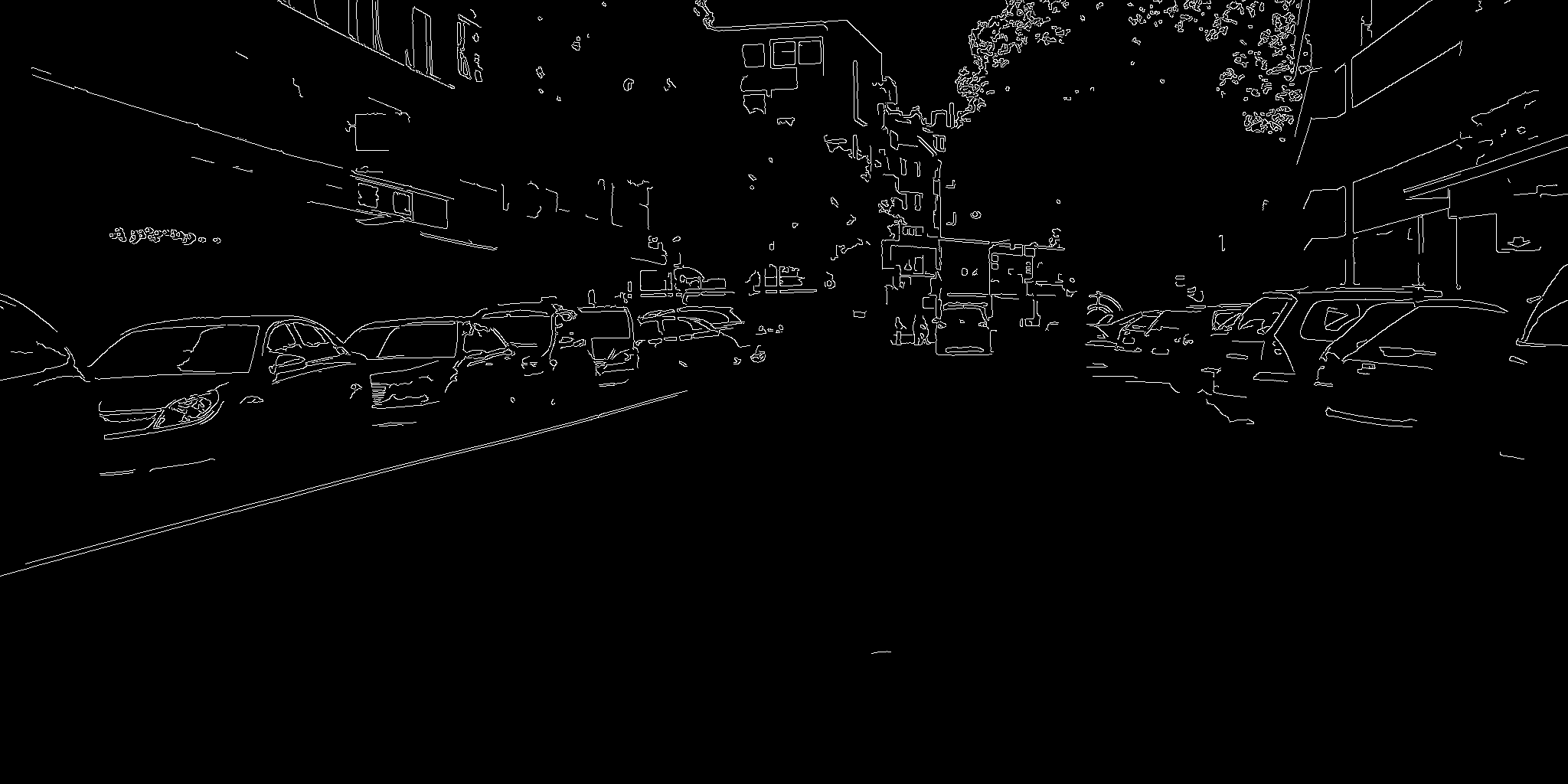}
\caption{Original image (top); Segmentation and edge maps (bottom): The segmentation map alone is not enough to distinguish different car structures; this structural detail is better-represented by edges.}
\label{fig:edges}
\end{figure*}

Given an input RGB image $I \in \mathbb{R}^{H \times W \times 3}$ that was captured by a traffic camera, the sender extracts three complementary semantic components, as depicted in Figure~\ref{fig:pipeline2}:
(i) a semantic segmentation layout, (ii) a structural edge representation, and (iii) a textual description of the scene.
Only these compact semantic modalities are transmitted to the other end over a communication channel.

Upon reception, a generative diffusion decoder conditions on the transmitted segmentation, edge map, and caption to synthesize the reconstructed image $\hat{I}$.
This allows for full semantic regeneration of the original scene without requiring transmission of pixel-level content.

The proposed framework follows an asymmetric sender-receiver design.
Sender-side computation consists of modules that can be deployed on resource-limited edge devices, such as a Raspberry Pi. At the same time, computationally intensive diffusion-based reconstruction is deferred to the receiver, where more powerful processing resources are available (e.g, GPU cards).

\subsection{Sender-Side Semantic Decomposition}

Once a picture is taken by the IoT device, the latter performs the following operations to decompose three components:\\
\textbf{Semantic Segmentation}: 
The input RGB image is processed using SegFormer~\cite{Xie2021SegFormerSA} to generate a semantic layout:
\[
S \in \{1, \ldots, C\}^{H \times W},
\]
where each pixel is assigned to one of $C$ semantic classes (e.g., vehicles, road, sky), while $H$ and $W$ are the dimensions of the image.
As camera images are usually of higher dimensions, the sender can reduce its size in order to also reduce the transmission size. Hence, the segmentation map is downsampled (e.g, to $1024 \times 512$ px) and compressed using WebP. WebP is a modern image format developed by Google that provides more efficient lossy compression than typical traditional formats, such as PNG and JPEG. In our case, it is particularly suitable for segmentation maps because these contain large uniform regions and a limited color palette, which WebP can encode compactly, while preserving the semantic layout.

This image representation is transmitted and later upsampled back to the original size during reconstruction (e.g., $2048 \times 1024$ in our experiments). 
We must note that despite the reduced resolution, the segmentation map preserves the spatial layout and object-level structure of the scene while remaining significantly smaller than the original RGB image. 

As shown in Figure~\ref{fig:edges}, the segmentation map alone may not provide detailed structure separation and spatial information. \\

\noindent \textbf{Edge Map}: 
To capture geometric boundaries and fine structural details, we extract an edge representation using the Canny edge detector. 
To do so, the RGB image is first converted to grayscale, after which Canny edge detection produces an edge map:
\[
E \in \{0, \ldots, 255\}^{H \times W}
\]
The resulting map (encoded as 0--255 intensity values) is compressed using WebP at the original resolution (e.g., $2048 \times 1024$) to preserve object contours without spatial downsampling. 
Compared to coarse pixel downsampling used in prior semantic codecs \cite{pezone2025semantic}, edge maps provide lightweight structural constraints that are inexpensive to compute and transmit (i.e., milliseconds of computation instead of seconds). 
\\

\begin{figure*}[!ht]
\centering
\includegraphics[width=520pt]{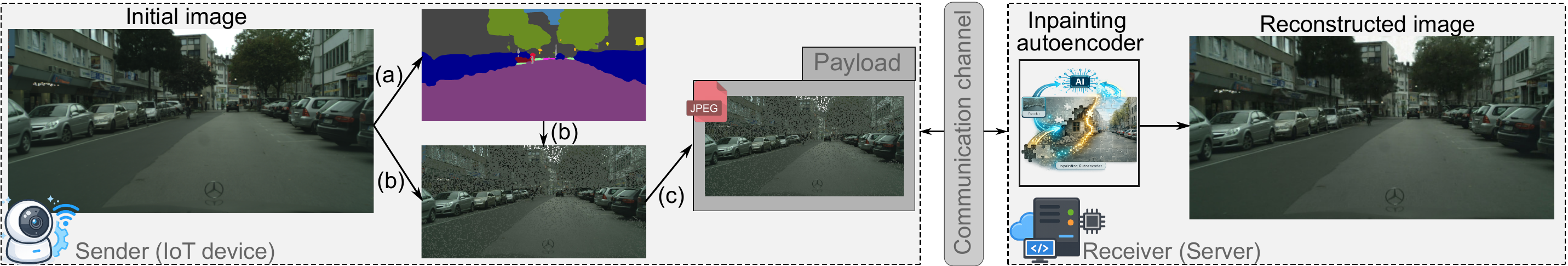}
\caption{Example of the semantic-aware masking and reconstruction pipeline (the class-of-interest is \textit{vehicles}); (a) the segmentation map is generated; (b) the masked image is produced, and (c) it is compressed prior transmission. The reconstructed image is produced at the receiver using the inpainting autoencoder.}
\label{fig:semantic_masking_example}
\end{figure*}

\noindent \textbf{Textual Description}: 
Finally, we generate a semantic textual description of the scene using a VLM, such as BLIP~\cite{li2022blipbootstrappinglanguageimagepretraining} or Gemma~3n~\cite{gemmateam2025gemma3technicalreport}. As shown in the evaluation, this textual information is needed as it encodes high-level contextual information about the objects and their relationships, complementing segmentation and edge maps, which capture finer structural details. 

Using a VLM model, we produce a caption:
\[
T = (w_1, \dots, w_L),
\]
consisting of $L$ words $w_i, i \in [1,L]$ that summarize high-level semantic content, including objects, activities, and contextual relationships present in the scene. $L$ is not predefined, but depends on the content of the image and the caption generated by the model. In practice, the caption remains short and compact, thereby contributing only a negligible overhead to the transmitted payload (i.e., some dozens of bytes).


\subsection{Receiver-Side Generative Reconstruction}

At the receiver, the transmitted semantic components $(S, E, T)$ are used as conditioning inputs to a diffusion-based generative decoder.
Prior to reconstruction, the segmentation map is upsampled to the target resolution (e.g., $2048 \times 1024$ px in our experiments), while the edge map is converted to an RGB-compatible representation suitable for conditioning the generative model.

The reconstruction process can be expressed as:
\[
\hat{I} = g_{\phi}(S, E, T),
\]
where $g_{\phi}$ denotes the conditional diffusion reconstruction model. 
The segmentation map constrains the spatial layout of objects, the edge map provides structural guidance for object boundaries, and the caption supplies high-level contextual information.

\section{SAMR: Semantic-Aware Image Compression with Masking and Generative Reconstruction}
\label{sec:masking}

This section presents SAMR, a semantic-aware image compression pipeline that selectively masks image regions prior to encoding, depending on their importance, thus, preserving the RoI structure. Semantic information is used to differentiate the treatment of image regions, allowing less relevant areas to be masked more aggressively while preserving critical content. The removed or heavily masked regions are reconstructed at the receiver using a learned inpainting model, enabling efficient transmission without requiring uniform compression across the entire image. 

Unlike MMSD, the focus of SAMR is to prioritize precise representation of RoI (e.g., vehicles or their plates), sacrificing data confidentiality, but overall maintaining a better quality/compression trade-off than existing approaches. More importantly, the two approaches are complementary as MMSD can serve as a lightweight first-stage representation, while SAMR can be invoked on demand when higher visual fidelity is required. Since SAMR relies on the same semantic segmentation maps as MMSD and computes masking with negligible overhead, a system operator can request a more detailed reconstruction without incurring significant additional processing cost at the edge, thus, enabling flexible and adaptive operation depending on the application requirements.

\subsection{Overall Pipeline}

Given an input image $I \in \mathbb{R}^{H \times W \times 3}$, the sender first extracts a semantic segmentation map and applies class-dependent masking at the patch level. The resulting masked image is then encoded using standard JPEG (or WebP) and transmitted. Upon reception, a learned inpainting autoencoder reconstructs the missing regions to obtain the final image $\hat{I}$.
Figure~\ref{fig:semantic_masking_example} illustrates the proposed pipeline. Starting from the original image, patches corresponding to semantically less relevant regions are selectively masked before JPEG encoding, yielding structured missing areas that are highly compressible. At the receiver, these regions are restored by the inpainting autoencoder, producing a visually coherent reconstruction that retains semantically important content.

As with the previous approach, this pipeline also follows the asymmetric sender--receiver design that is well suited for deployment in edge environments. Processing on the sender side is limited to lightweight operations, including segmentation, masking, and standard JPEG or WebP encoding, which can be efficiently executed on low-end microprocessors. In contrast, computationally intensive reconstruction is deferred to the receiver. 

\subsection{Sender-Side Processing}

Once a camera picture is taken, the sender executes three lightweight operations to reduce the data transfer size:\\
\noindent \textbf{Semantic Segmentation}: This is the same operation presented in the previous approach, where the input image is processed using SegFormer, assigning each pixel to one of $C$ semantic classes (e.g., vehicle, road, sky, vegetation). However, in this approach, we go one step further and we associate each class $c$ with a predefined masking ratio $\rho_c \in [0,1]$, reflecting its relative importance. This assignment allows us to prioritize semantic classes differently by dropping out image patches more or less aggressively for certain classes. \\

\noindent \textbf{Semantic-Aware Patch Masking}: The image is partitioned into non-overlapping patches of size $8 \times 8$ pixels. For each patch, the dominant semantic class is identified by majority voting over the corresponding region in $S$. Let $c_{ij}$ denote this class. A binary mask value $M_{ij}$ is sampled as:
\[
M_{ij} \sim \text{Bernoulli}(\rho_{c_{ij}}),
\]
where $M_{ij}=1$ indicates that the patch is removed from the image. Masked patches are replaced with zero-valued pixels, producing a partially observed image $\tilde{I}$. Patch-level masking results in spatially coherent missing regions that facilitate generative inpainting.\\

\noindent \textbf{JPEG/WebP Encoding}: The masked image $\tilde{I}$ is encoded using standard JPEG or WebP at a configurable quality factor $Q$. The quality factor serves as a secondary compression knob: lower values of $Q$ produce smaller files at the cost of introducing compression artifacts into the preserved image regions. The presence of large zero-valued regions from masking increases spatial redundancy, yielding improved compression efficiency compared to encoding the original image at the same $Q$. Unlike the previous approach, only the JPEG or WebP masked bitstream is transmitted to the receiver for reconstruction.

\subsection{Receiver-Side Reconstruction}

After decoding the JPEG/WebP bitstream, the receiver recovers the masked image $\tilde{I}$. A learned inpainting autoencoder $f_\theta$ reconstructs the missing regions according to
$\hat{I} = f_\theta(\tilde{I})$, 
using the visible context to infer plausible content for the removed patches. 
In our framework, masking is not applied uniformly, but is guided by semantic importance so that non-critical image patches are selectively removed prior to standard JPEG/WebP encoding, while semantically important regions are more likely to be preserved. This semantic-aware masking strategy enables substantial bitrate reduction at the sender, but retains sufficient contextual and structural information for the receiver-side generative inpainting model to plausibly reconstruct the missing content and preserve the overall semantic consistency of the scene. 

\subsection{Masking Strategies}
\label{sec:patch-based}

SAMR employs two masking strategies with distinct purposes. During offline training, we use semantic-agnostic random masking to improve the robustness and generalization of the inpainting model. During inference, however, we switch to semantic-aware masking, which prioritizes important content and improves compressibility. This separation ensures that the reconstruction model learns a strong generative prior while the deployed system benefits from structured, task-aware masking.

\begin{figure}
    \centering
    \includegraphics[width=280pt]{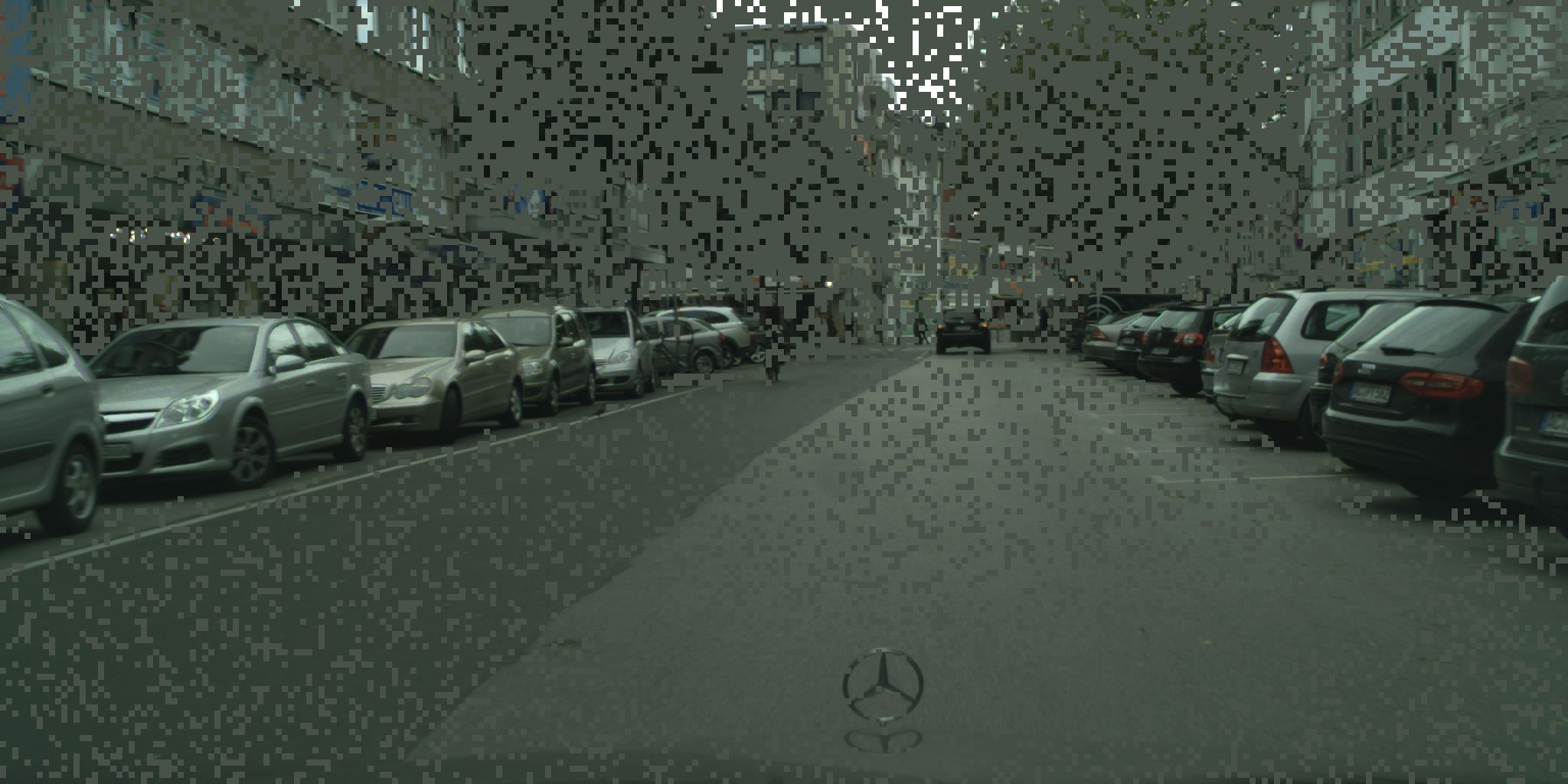}
    \caption{Example of Semantic-Aware Masking in SAMR: RoI (i.e., vehicles) are not masked at all or are lightly masked, but non-RoI (i.e., road, buildings, and sky) are heavily masked.}
    \label{fig:masking}
\end{figure}

\subsubsection{Random Masking}
\label{sec:random-masking}

Given an input image $I \in \mathbb{R}^{H \times W \times 3}$, we divide it into non-overlapping patches of size $8 \times 8$ px, producing a grid of $n_h \times n_w$ patches, where $n_h = H/8$ and $n_w = W/8$. 

For a specified masking ratio $\rho \in [0, 1]$, we randomly select $\lfloor \rho \cdot n_h \cdot n_w \rfloor$ patches and replace them with zeros. The resulting binary patch mask $M \in \{0,1\}^{n_h \times n_w}$ indicates masked ($M_{ij}=1$) and preserved ($M_{ij}=0$) patches. 

Random masking treats all regions uniformly, independent of semantic content. This forces the inpainting autoencoder to reconstruct arbitrary missing regions and prevents overfitting to specific masking patterns. As a result, the model learns more general contextual representations and becomes robust to varying mask geometries and densities.

\subsubsection{Semantic Importance-Aware Masking}
\label{sec:hierarchical-masking}

During inference, masking probabilities are modulated according to semantic importance, enabling application-driven prioritization of visual content. Semantic classes are grouped into coarse categories based on their relative relevance to the target task, and each category is assigned a masking probability $\rho_c$, allowing importance policies to be adjusted without modifying the compression pipeline.

\begin{figure*}
    \centering
    \subfloat[Cityscapes dataset~\cite{Cordts2016Cityscapes}.]{\includegraphics[height=120pt]{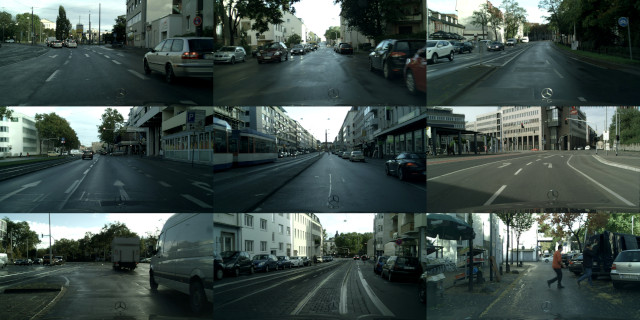}}\hspace{2pt}
    \subfloat[Sochor et al. dataset \cite{sochor2018comprehensive}.]{\includegraphics[height=120pt]{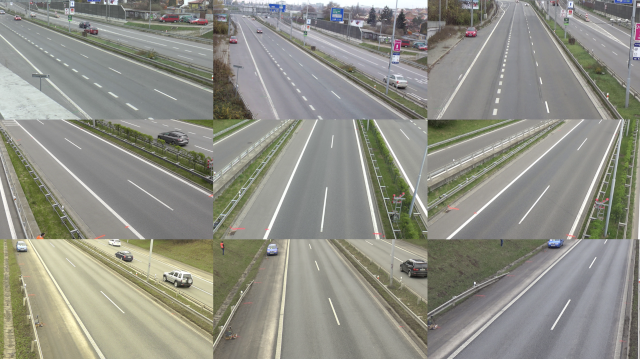}}
    \caption{Samples of images/frames of the two employed datasets.}
    \label{fig:datasets}
\end{figure*}

As mentioned earlier, masking is applied at the patch level by first identifying the dominant semantic class $c_{ij}$ of each patch and then sampling the corresponding mask value as $M_{ij} \sim \text{Bernoulli}(\rho_{c_{ij}})$. As shown in the example of Figure~\ref{fig:masking}, unlike other approaches in the literature, this procedure yields masks that are aligned with semantic regions rather than being spatially uniform.

To support operation under different bandwidth constraints, multiple importance configurations are defined by varying $\rho_c$ across semantic categories. Lower configurations emphasize preservation of high-importance content, while higher configurations increase compression by progressively masking a larger portion of the image. The impact of these configurations is assessed in Section \ref{sec:eval}.

\subsection{Inpainting Autoencoder and Training Procedure}
\label{sec:inpainting-ae}

The masked regions are reconstructed at the receiver using a convolutional inpainting autoencoder. The network follows an encoder–decoder architecture with skip connections to preserve spatial detail.

The encoder progressively downsamples the masked input image $\tilde{I}$ to extract hierarchical feature representations. Four residual blocks at the bottleneck refine the latent representation and increase receptive field capacity. The decoder then progressively upsamples the features back to the original resolution, incorporating skip connections from corresponding encoder layers to retain fine-grained spatial information. The network outputs a reconstructed RGB image $\hat{I} \in \mathbb{R}^{H \times W \times 3}$ without additional output activation.

The network is trained using a composite loss function that combines a mask-weighted $L_1$ loss with a VGG-based (Visual Geometry Group) perceptual loss~\cite{johnson2016perceptuallossesrealtimestyle}. $L_1$ assigns higher weight to masked regions (e.g., $0.7$) than to unmasked regions (e.g., $0.3$), encouraging the model to focus reconstruction effort where content was removed. The perceptual loss computes the $L_1$ distance between feature representations extracted from intermediate layers of a pretrained VGG-16 network. To reduce memory overhead on high-resolution inputs, both the predicted and target images are downscaled (e.g., to $256 \times 512$ px) before feature extraction. The total loss is defined as:
\[
\mathcal{L} = \mathcal{L}_{L_1}^{\text{weighted}} + \lambda_p \mathcal{L}_{\text{perceptual}},
\]
where $\lambda_p$ controls the weight of the perceptual term. This combination encourages both pixel-level accuracy and perceptually coherent texture reconstruction. 


\section{Evaluation \& Discussion of the Results}
\label{sec:eval}

We evaluated the two semantic compression strategies along two dimensions: reconstruction fidelity - semantic preservation, and compression rate. Rather than directly comparing the two approaches, we evaluate each of them under their intended operational setting with approaches in the literature.

\subsection{Datasets and Experiments Setup}

\subsubsection{Datasets}

We assessed the proposed frameworks on two datasets. The Cityscapes dataset~\cite{Cordts2016Cityscapes}, which consists of high-resolution urban street scenes captured from vehicle-mounted cameras and includes diverse road layouts, vehicles, pedestrians, and surrounding infrastructure. We also employed the dataset from Sochor et al. \cite{sochor2018comprehensive}--which consists of videos from traffic control cameras. From the datasets, we randomly sampled 500 images/frames for testing purposes and used another 3,000 images for training. Examples of images and frames of these datasets are depicted in Figure~\ref{fig:datasets}.

\subsubsection{Setup for MMSD}

To evaluate MMSD, we extracted the three semantic components, as described in Section~\ref{sec:multimodal}: semantic segmentation maps, edge maps, and textual descriptions. 
For each image, we generated one semantic segmentation map and one edge map, resulting in 500 segmentation maps and 500 edge maps in total.
The edge map was generated using the Canny edge detector with the default threshold settings, requiring no manual tuning. 
To obtain textual representations, we generated two captions per image using BLIP and Gemma, yielding 1,000 captions overall. We employed two different caption generation models, as Gemma tends to produce longer and more descriptive captions compared to BLIP. Thus, we wanted to see how this affects the reconstruction quality.

At the receiver side, each caption was independently combined with the corresponding segmentation and edge information to reconstruct the image, producing 1,000 reconstructed images in total. To incorporate multiple structural constraints simultaneously, we used the pretrained Stable Diffusion v1.5 model architecture~\cite{Zhang_2023_ICCV}, which helps diffusion models to be guided by several control signals in parallel. Then we connected two distinct ControlNet models into the main one to condition the generation on our two spatial modalities (segmentation and edge maps). The first model is called ControlNet-Seg, and the second is ControlNet-Canny. The textual caption is used as the text prompt. ControlNet‑Seg receives the upsampled and resized segmentation map, while ControlNet‑Canny receives the resized edge map (converted to a three‑channel image). The text caption \(T\) is encoded using the CLIP text encoder that was trained with Stable Diffusion v1.5. It produces a sequence of embeddings that serve as the prompt for the diffusion model. All three components together are used to produce the reconstructed image.   

We compare MMSD to one of the most recent studies in the literature, named SPIC \cite{pezone2025semantic}. SPIC employs segmentation without any additional elements such as edges or captions. Unlike MMSD, the segmentation map is not downscaled, in order to preserve as much spatial information as possible. 

\subsubsection{Setup for SAMR}

\begin{table}
\centering
\caption{Hierarchical masking in SAMR: class-specific masking probabilities for selected configurations.}
\label{tab:hierarchical-configs}
\setlength\tabcolsep{2pt}
\small
\begin{tabular}{lcccc}
\toprule
\textbf{Semantic} & \textbf{Config 0} & \textbf{Config 2} & \textbf{Config 4} & \textbf{Config 7} \\
\textbf{ Group} & {\footnotesize (Conservative)} & {\footnotesize (Moderate)} & {\footnotesize (Balanced)} & {\footnotesize (Aggressive)}\\
\midrule
Vehicles     & 0.0 & 0.2 & 0.4 & 0.4 \\
Humans       & 0.2 & 0.4 & 0.5 & 0.6 \\
Flat surfaces & 0.2 & 0.4 & 0.5 & 0.6 \\
Construction & 0.5 & 0.6 & 0.7 & 0.8 \\
Objects      & 0.5 & 0.6 & 0.7 & 0.8 \\
Nature       & 0.8 & 0.8 & 0.8 & 0.9 \\
Sky          & 0.8 & 0.8 & 0.8 & 0.9 \\
Background   & 0.8 & 0.8 & 0.8 & 0.9 \\
\bottomrule
\end{tabular}
\end{table}

Regarding SAMR, we employed 8 configuration classes, as shown in Table~\ref{tab:hierarchical-configs}, to assess the aggressiveness of semantic-aware masking across different segment classes. These configurations progressively vary the masking probabilities assigned to each semantic category, enabling evaluation under different compression levels. In addition, we varied the JPEG quality factor over a wide range ($Q \in [5, 100]$) to analyze the interaction between masking strength and conventional compression. This combined setup allows us to explore a broad spectrum of operating points and characterize the trade-off between compression efficiency and reconstruction quality. JPEG was employed instead of WebP to directly compare SAMR with conventional and widely adopted image compression pipelines, thereby isolating the gain introduced by semantic-aware masking and receiver-side reconstruction rather than by differences between codecs.

At the receiver, we used the AdamW optimizer with learning rate $\eta = 5 \times 10^{-4}$, weight decay $\lambda = 0.01$, and $\lambda_p = 0.1$.  Training was performed for 20 epochs with batch size 8. The model was trained with data-parallel distribution across two NVIDIA T4 GPUs (Kaggle platform), requiring approximately 30 hours.

During training, we applied two forms of randomized augmentation. First, semantic-agnostic random masking was used rather than the semantic-aware masking employed at inference time. For each image and training iteration, a new masking pattern was generated with masking ratio $\rho \sim \text{Uniform}(0.1, 0.8)$. Second, each masked image was subjected to a random JPEG compression round-trip at quality $Q \sim \text{Uniform}(10, 80)$, simulating the variable JPEG encoding applied during inference. This dual randomization strategy prevents overfitting to any specific masking structure or artifact pattern and encourages the network to learn a robust image completion prior. Although inference uses deterministic semantic-aware masking, the randomized training distribution sufficiently covers the variability encountered across different operating points.

For comparative analysis, we selected two baseline codecs: the standard JPEG codec \cite{wallace1992jpeg} and the recent neural codec SQ-GAN \cite{Pezone2025sqgan}. SQ-GAN was selected due to the fact that it uses a semantic masking concept similar to SAMR, with the difference being the application method. While our approach uses masking as a preprocessing step before the actual encoder, SQ-GAN incorporates masking directly into the rate-distortion optimization process of the encoder. 

\subsection{Evaluation of the MMSD Framework}

This subsection evaluates the effectiveness of MMSD in terms of compression efficiency and semantic preservation. We demonstrate that decomposing images into segmentation maps, edge representations, and compact textual captions enables substantial data reduction to the original image, while better maintaining high-level structural and contextual information necessary for faithful reconstruction compared to SPIC.

\subsubsection{Compression Efficiency}

\begin{table}
\centering
\caption{Payload Transfer Size Comparison between MMSD and SPIC (all values in KBytes, except captions).}
\label{tab:payload}
\setlength\tabcolsep{3.5pt}
\small
\begin{tabular}{l cc cc}
\hline
\textbf{Component} & \multicolumn{2}{c}{\textbf{MMSD}} & \multicolumn{2}{c}{\textbf{SPIC}} \\
\cline{2-3} \cline{4-5}
 & \textbf{Mean} & \textbf{$\sigma$} & \textbf{Mean} & \textbf{$\sigma$} \\
\hline
Segmentation Map & 3.96 & 0.93  & 12.74 & 2.58 \\
Edge Map & 18.41 & 6.79 & -- & -- \\
Caption (BLIP) & 57\,B & 5.02\,B & -- & -- \\
Caption (Gemma) & 260.49\,B & 26.59\,B & -- & -- \\
Coarse Image & -- & -- & 10.60 & 1.67 \\
\hline
\textbf{Total Payload Size} & 22.42 & 6.94& 23.34 & 3.37 \\
\hline
\textbf{Compression Ratio} & 111.12$\times$ & 40.03 & 99.91$\times$ & 11.50 \\
\hline
\end{tabular}
\end{table}

Table~\ref{tab:payload} demonstrates a substantial reduction in transmitted data achieved by the proposed semantic representation compared to the original RGB image. While the average size of the original image is 2.28\,MB, the segmentation map and edge map require only 3.96\,KB and 18.41\,KB, respectively. The textual captions contribute a negligible overhead, with the BLIP caption averaging just 57\,B and the Gemma caption 261\,B, both accounting for less than 0.02\% of the original image size. Overall, the complete transmitted payload, including segmentation, edge map, and caption, amounts to approximately 22.42\,KB, corresponding to only 0.96-0.98\% of the original RGB image. 
Compared to SPIC, the average payload size of MMSD is also slightly lower.

\subsubsection{Human Perceptual Evaluation of the Reconstruction}

To assess perceptual reconstruction quality, we conducted a human evaluation study and a VLM-based evaluation. In the first one, participants compared original images to their reconstructed counterparts.
Each participant was presented with 20 randomly selected image triplets that consisted of the original image and two reconstructed images: one generated by the SPIC baseline and one generated by MMSD (using either BLIP or Gemma captions). The original image was shown as a reference, while participants compared the two reconstructions relative to it. 

For each triplet, participants answered two questions which evaluated semantic preservation in the reconstructions:\\
(Q1) whether the spatial positions of vehicles were preserved, and \\
(Q2) whether the number of vehicles was preserved.

The questionnaire was structured as a forced-choice comparison between the two reconstructions. 
For each question, participants could select one of four options: the first reconstruction, the second reconstruction, both reconstructions (if they preserved the property equally well), or none (if neither reconstruction preserved the property sufficiently). This triplet-based evaluation was used for direct pairwise comparison between reconstruction methods while ensuring that both reconstructions are judged under identical visual context. Full statistics of the survey are shown in Table~\ref{tab:survey_setup}.
In total, 51 participants made 1238 responses across randomly assigned image triplets.

\begin{table}
\centering
\caption{Human Survey Setup and Coverage (MMSD vs SPIC).}
\label{tab:survey_setup}
\setlength\tabcolsep{3pt}
\small
\begin{tabular}{l r}
\toprule
\textbf{Metric} & \textbf{Value} \\
\midrule
Total participants & 51 \\
Total reconstructed image triplets & 500 \\
Unique image pairs shown & 343 \\
Pairs per participant & 20 \\
Pairs with at least one response & 343 \\
Total responses collected & 1,238 \\
\bottomrule
\end{tabular}
\end{table}

We let participants compare two reconstructions within each triplet to derive pairwise preferences among two methods. The cases when either both or none methods were able to satisfy the questions criteria we considered as ties. Ties occurred in 49.1\% of comparisons for positions preservation (Q1) and in 39.8\% of comparisons for vehicles count preservation (Q2). When we were computing the preference rates, we excluded those ties and used only decisive responses.

\begin{table}
\centering
\caption{Human survey: Preference against the SPIC baseline.}
\label{tab:human_spic}
\small
\setlength\tabcolsep{4pt}
\begin{tabular}{l c c}
\toprule
\textbf{Metric} & \textbf{MMSD} & \textbf{SPIC} \\
\midrule
Position preservation (Q1) & \textbf{63.2\%} & 36.8\% \\
Number preservation (Q2)   & \textbf{68.4\%} & 31.6\% \\
\midrule
\textbf{Overall win rate}  & \textbf{66.0\%} & 34.0\% \\
\bottomrule
\end{tabular}
\\[0.4em]
\footnotesize{Results computed on decisive responses only. MMSD significantly outperforms SPIC ($p < 0.001$, binomial test).}
\end{table}

\begin{figure*}[t]
\centering
\subfloat[Original]{\includegraphics[width=0.32\textwidth]{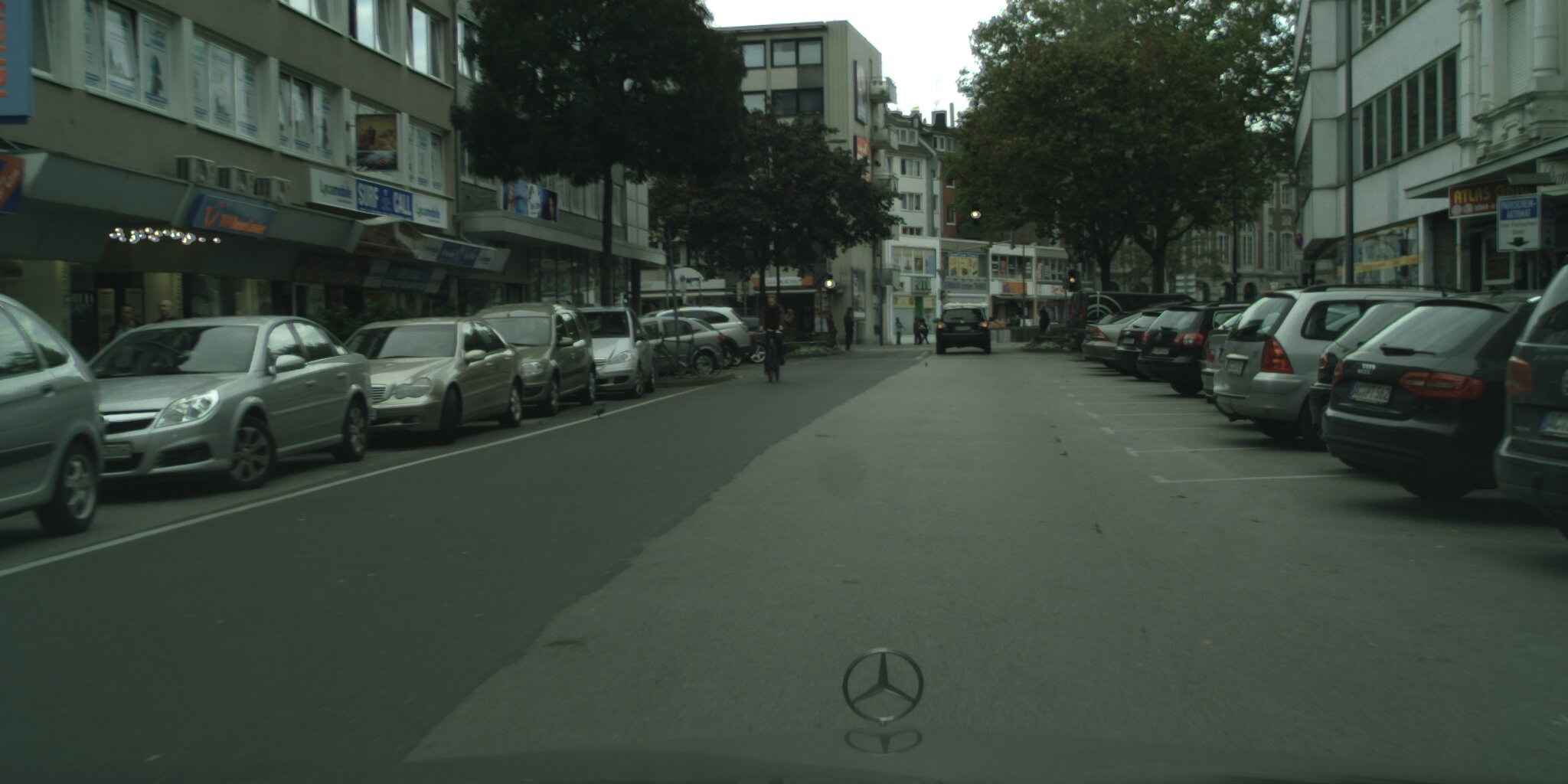}}\hfill
\subfloat[SPIC]{\includegraphics[width=0.32\textwidth]{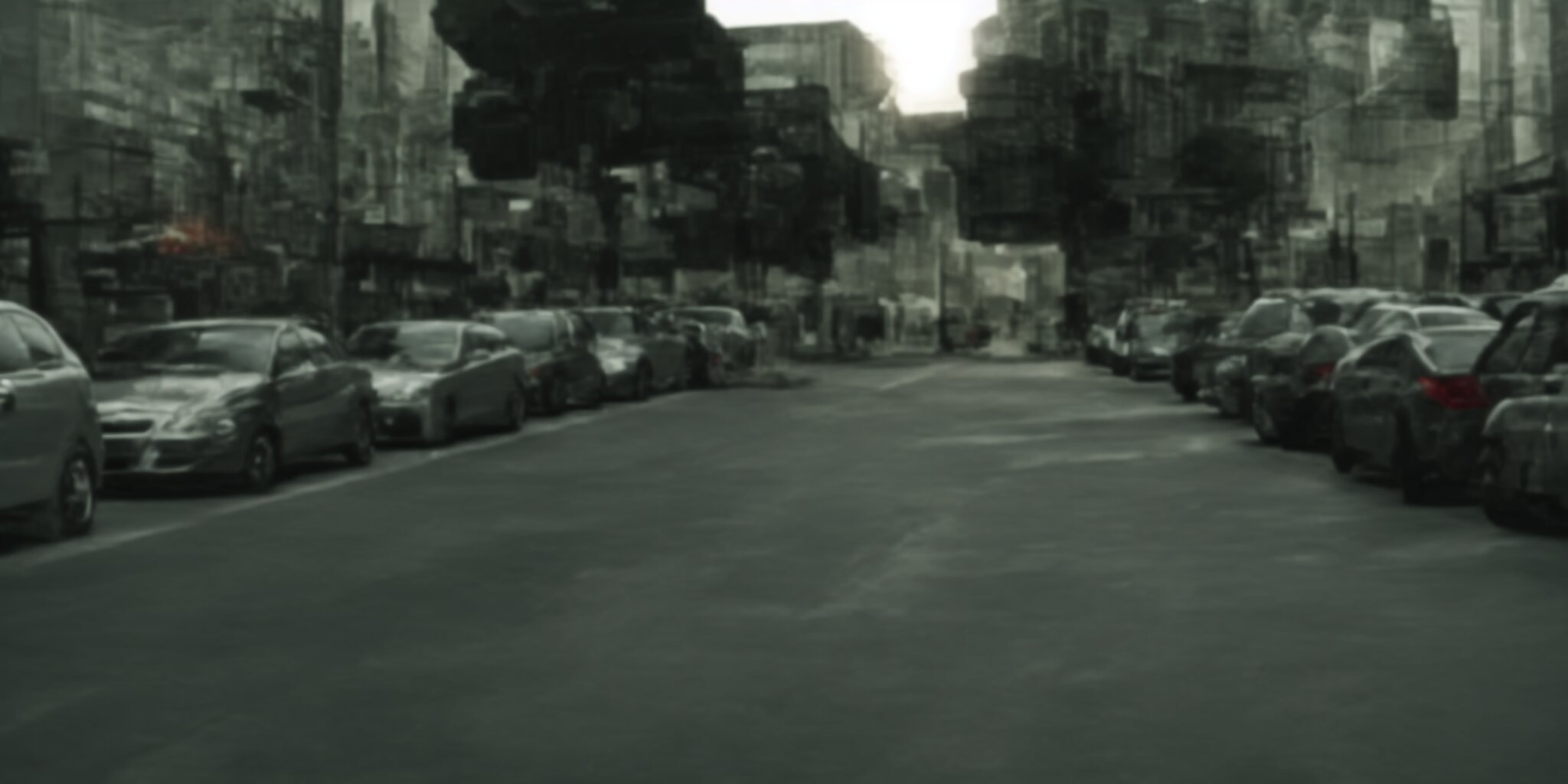}}\hfill
\subfloat[MMSD with Gemma caption]{\includegraphics[width=0.32\textwidth]{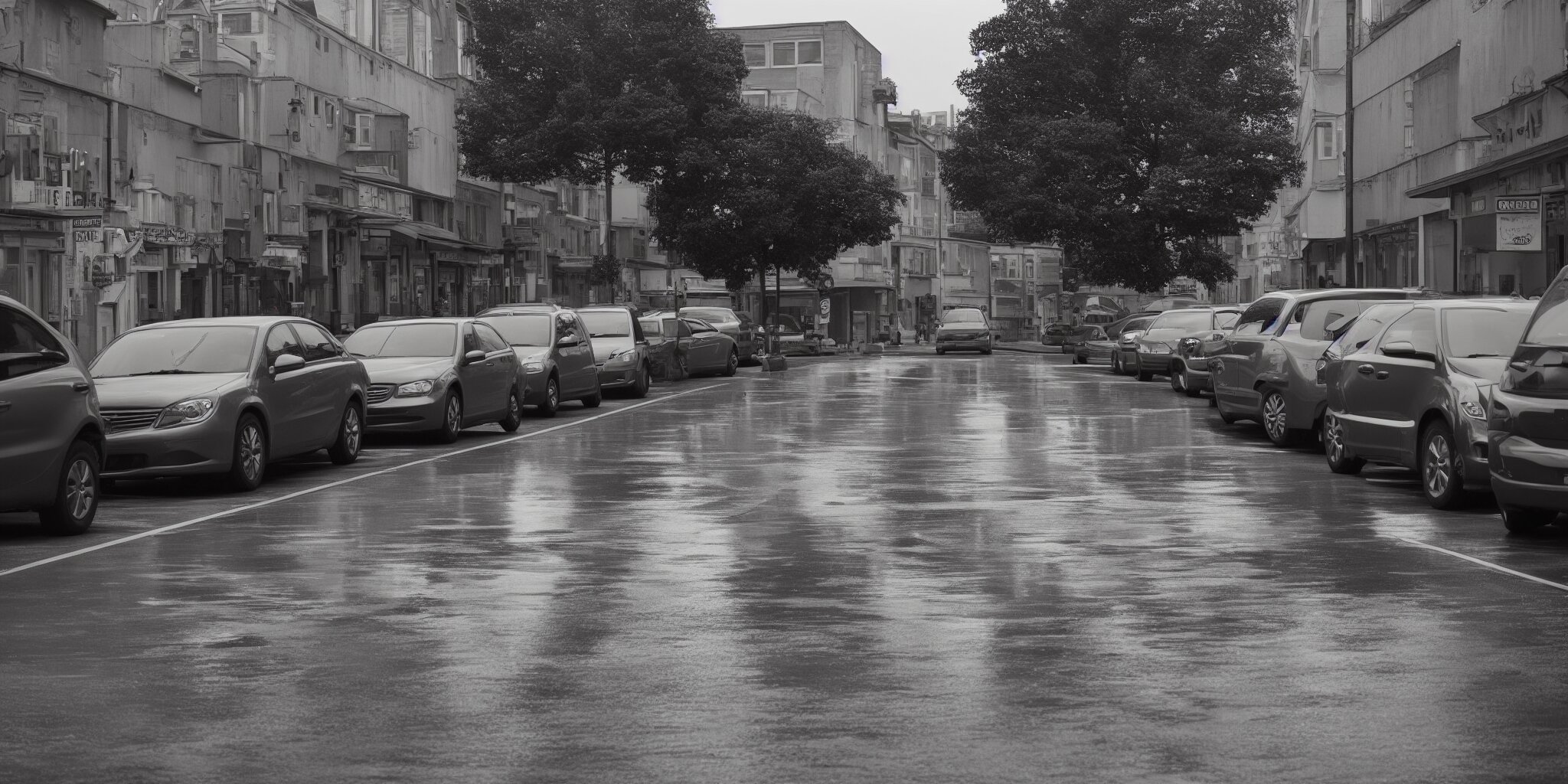}}
\vspace{5pt}
\subfloat[Original (crop)]{\includegraphics[width=0.32\textwidth]{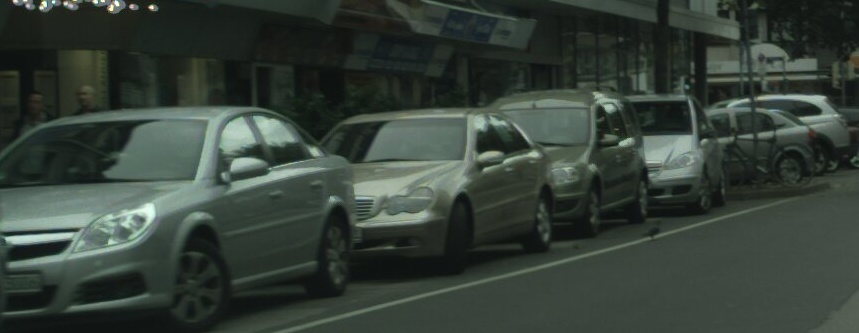}}\hfill
\subfloat[SPIC (crop)]{\includegraphics[width=0.32\textwidth]{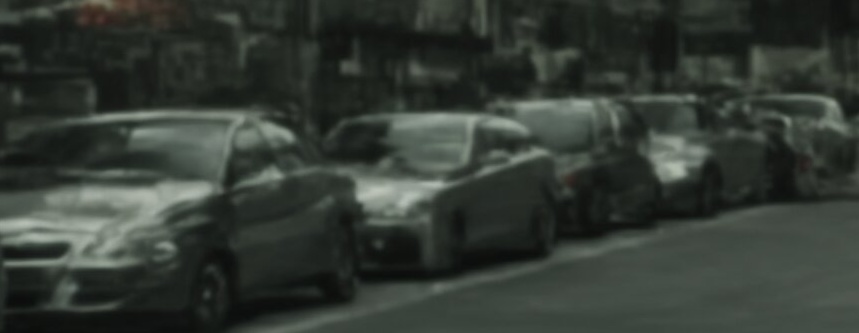}}\hfill
\subfloat[MMSD (crop)]{\includegraphics[width=0.32\textwidth]{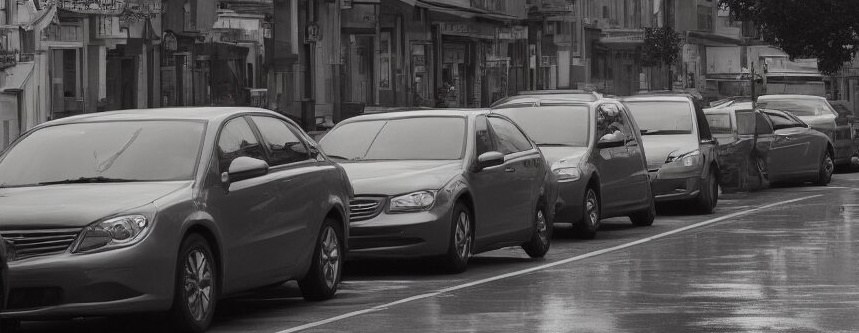}}
\caption{Qualitative comparison between the original image and reconstructed ones (MMSD vs SPIC). Top row: full images; Bottom row: zoomed-in crops ($859 \times 333$ px) highlighting reconstruction details.}
\label{fig:caption_comparison_good}
\end{figure*}

Table~\ref{tab:human_spic} shows the pairwise preference rates that were computed only from decisive responses. The proposed method is preferred over the SPIC baseline in 63.2\% of comparisons for position preservation and in 68.4\% of comparisons for vehicle count preservation. These results indicate that the tri-modal semantic representation provides more reliable preservation of semantic information in reconstructed traffic scenes compared to SPIC. Qualitatively, reconstructions that were produced by SPIC often exhibit blurred vehicle regions where object boundaries become indistinguishable, as shown in Figure~\ref{fig:caption_comparison_good}. 
In contrast, the proposed method frequently produces sharper object structures that are easier for participants to interpret. 
This improvement can be attributed to the combination of semantic segmentation and edge maps: the segmentation map constrains the spatial layout of objects, while the edge map provides additional structural cues that help the generative decoder better preserve object contours.

We also analyze the effect of the caption models used in the proposed pipeline. 
Table~\ref{tab:vlm_compare} compares the two captioning models that were  used in our method. Both variants outperform the SPIC baseline, where the BLIP variant achieved a preference rate of 64.6\% and the Gemma variant achieved 67.5\%.


\begin{table}[t]
\centering
\caption{Human survey: Win rate of the proposed method over SPIC by caption model (decisive responses only).}
\label{tab:vlm_compare}
\small
\setlength\tabcolsep{4pt}
\begin{tabular}{l c c}
\toprule
\textbf{Metric} & \textbf{BLIP} & \textbf{Gemma} \\
\midrule
Win rate over SPIC & 64.6\% & \textbf{67.5\%} \\
\bottomrule
\end{tabular}
\end{table}

\begin{figure*}
\centering
\subfloat[Original image]{\includegraphics[width=0.32\linewidth]{img/aachen_000013_000019.jpg} \label{fig:orig}}\hspace{2pt}
\subfloat[Reconstruction using the BLIP-generated caption: ``\textit{There are many cars parked on the side of the street}''.]{\includegraphics[width=0.32\linewidth]{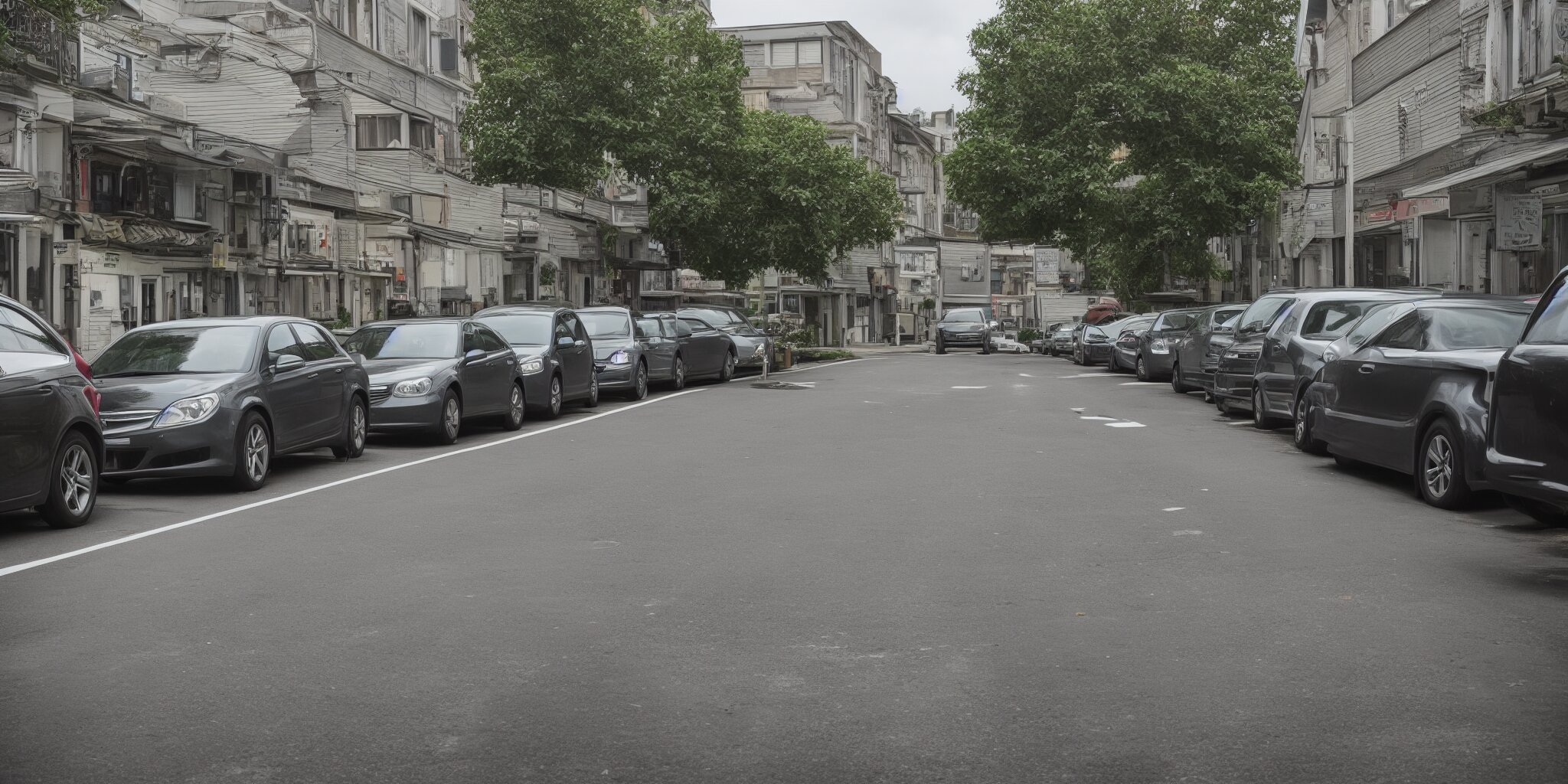} \label{fig:blip}}\hspace{3pt}
\subfloat[Reconstruction using the Gemma-generated caption: ``\textit{A quiet city street scene shows parked cars lining both sides of the road. Buildings with shops and trees line the street, with a few pedestrians visible. The sky is overcast, and the overall atmosphere is calm and urban}''.]{\includegraphics[width=0.32\linewidth]{img/good_example_damir/aachen_000013_000019_gemma_generated_comp5.jpg}\label{fig:gemma}}
\caption{Qualitative comparison between the original image and MMSD's reconstructions guided by BLIP and Gemma captions.}
\label{fig:caption_comparison}
\end{figure*}

Although BLIP produces shorter textual descriptions and relies on a significantly lighter VLM, it achieves slightly higher perceptual preference rates. 
In contrast, Gemma generates longer and more descriptive captions (see the example in Figure~\ref{fig:caption_comparison}) using a much larger model, yet produces comparable reconstruction quality.

\footnote{The human evaluation survey was conducted using Qualtrics and is available online at: \url{https://nukz.qualtrics.com/jfe/form/SV_2oyeGkhBLc5Mboa}.}

Overall, the human evaluation indicates that reconstructed images preserve object positions with high fidelity. At the same time, preservation of the exact number of vehicles remains more challenging. 
The results also suggest that compact captions that are generated by lightweight models such as BLIP are sufficient to guide the reconstruction process at the same quality as larger models such as Gemma.

\subsubsection{VLM-based Evaluation of the Reconstruction}

Following prior work on VLM-based evaluation of multimodal outputs~\cite{wang2024large,Han_2026,sharshar2025vision}, and in addition to human evaluation we employed a VLM, more specifically Gemini 27B, as a judge to assess semantic preservation in reconstructed images. 
The VLM was presented with pairs of images that consisted of the original image and a reconstructed image, and was asked to assess whether key semantic properties are preserved.
Specifically, we evaluated all reconstructed images generated in our experiments.
For each reconstruction, the VLM received either an [original, reconstructed\_BLIP] or an [original, reconstructed\_Gemma] image pair. 
The evaluation criteria mirror those used in the human study, focusing on object position preservation and object count preservation.
For each image pair, the VLM produces binary judgments for both criteria, enabling direct comparison with human perceptual results. To ensure consistency across evaluations, we use a fixed prompt and deterministic decoding settings for all VLM judgments.

\begin{table}[t]
\centering
\caption{VLM-based Evaluation Results (Gemini 27B).}
\label{tab:llm_judge}
\setlength\tabcolsep{5pt}
\small
\begin{tabular}{l c c c}
\toprule
\multirow{2}{*}{\textbf{Metric}} & \multicolumn{2}{c}{\textbf{MMSD}} & \multirow{2}{*}{\textbf{SPIC}} \\
\cline{2-3}
& \textbf{BLIP} & \textbf{Gemma} & \\
\midrule
Position preservation (Q1) & 94.44 & \textbf{97.61} & 93.65 \\
Number preservation (Q2)   & \textbf{88.89} & 86.90 & 80.16 \\
Joint success rate (Q1 \& Q2) & \textbf{86.9} & 86.06 & 75.07 \\
\bottomrule
\end{tabular}
\end{table}

\begin{table}[t]
\centering
\caption{Ablation study of MMSD (Gemini 27B).}
\label{tab:llm_ablation}
\setlength\tabcolsep{3pt}
\small
\begin{tabular}{l cc cc c}
\toprule
\multirow{2}{*}{\textbf{Metric}} 
& \multicolumn{2}{c}{\textbf{Full pipeline}} 
& \multicolumn{2}{c}{\textbf{No edge map}} 
& \textbf{No} \\
\cline{2-3} \cline{4-5}
& \textbf{BLIP} & \textbf{Gemma} & \textbf{BLIP} & \textbf{Gemma} & \textbf{caption}\\
\midrule
Position (Q1) & 94.44 & \textbf{97.61} & 52.38 & 58.73 & 76.49 \\
Number (Q2)   & \textbf{88.89} & 86.90 & 28.57 & 34.13 & 54.58 \\
Joint         & \textbf{86.90} & 86.06 & 24.21 & 28.17 & 50.8 \\
\bottomrule
\end{tabular}
\end{table}

\begin{figure*}
\centering
\subfloat[Segmap + Edge map + Caption]{\includegraphics[width=153pt]{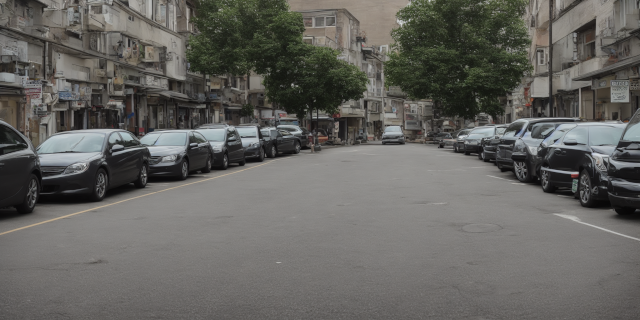}}\hspace{1pt}
\subfloat[Segmap + Caption]{\includegraphics[width=153pt]{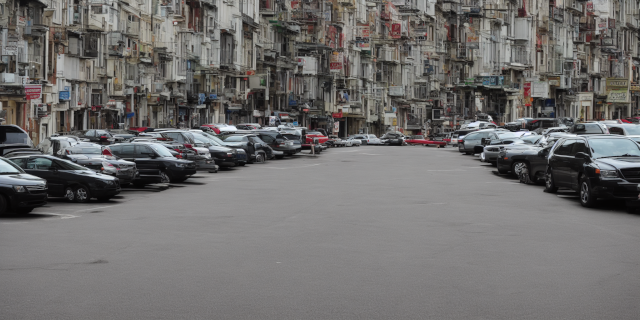}}\hspace{1pt}
\subfloat[Segmap + Edge map]{\includegraphics[width=153pt]{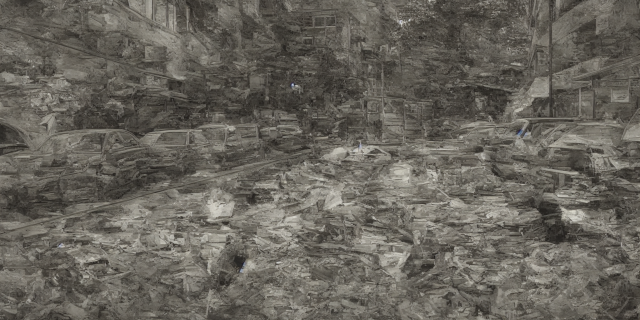}}
\caption{Qualitative comparison of MMSD reconstruction with (a) the full pipeline (BLIP) and (b-c) combinations of missing transferred elements.}
\label{fig:without_edges}
\end{figure*}

Table~\ref{tab:llm_judge} presents the VLM-based semantic evaluation results using Gemini 27B as an automated judge, measuring how well reconstructed images preserve object positions and counts. 
Both variants of the proposed method outperform the SPIC baseline across all evaluated metrics. 
For position preservation (Q1), the Gemma-based configuration achieves the highest score at 97.61\%, followed by BLIP at 94.44\%, while SPIC reaches 93.65\%. 
For number preservation (Q2), the BLIP configuration slightly outperforms Gemma, achieving 88.89\% compared to 86.90\%, while SPIC achieves 80.16\%. 
A similar trend is observed for the joint success rate (Q1 \& Q2), where BLIP achieves 86.9\%, Gemma achieves 86.06\%, and SPIC reaches 75.07\%. 

\subsubsection{Ablation Study of MMSD}

To see the impact of each MMSD component in relation to the reconstructed image, we performed an ablation study. Specifically, we excluded the edge map and caption from the payload to see which component affects the most the reconstruction quality. We decided not to include the segmentation map in the study as it is the main component of the reconstruction. 

Table \ref{tab:llm_ablation} shows that the configuration without the edge map performs poorly on both caption options. There is also a trend that Gemma-generated captions tend to perform better when there is no edge map as it portrays more details of the original scene. The configuration without the caption produced better results, but still lower than the full pipeline. The effect of the two configurations on the reconstructed image is also illustrated in the example of Figure~\ref{fig:without_edges}. In this example, while the absence of the edge map produces a comprehensible scene reconstruction, the absence of the caption leads to a catastrophic representation.


\subsubsection{Edge Deployment \& Inference Latency}

Finally, to evaluate deployment feasibility on resource-constrained devices, we measured the inference latency for each pipeline component on a Raspberry Pi~5 using CPU-only inference.

\begin{table}[t]
\centering
\caption{Inference time and memory usage on a Raspberry Pi~5.}
\label{tab:rpi_MMSD}
\setlength\tabcolsep{4pt}
\small
\begin{tabular}{l c c c c}
\toprule
\textbf{Metric} & \textbf{Segmentation} & \textbf{Edge} & \textbf{Caption} & \textbf{Total} \\
\midrule
\multicolumn{5}{c}{\textbf{MMSD}} \\
Time (s)      & 8.35 & 0.02 & 7.07 & 15.46 \\
Memory (MB)   & $\sim$400 & $<$10 & $\sim$1,000 & $\sim$1,400 \\
\midrule
\multicolumn{5}{c}{\textbf{SPIC}} \\
Time (s)      & 8.35 & --- & --- & 8.35 \\
Memory (MB)   & $\sim$400 & --- & --- & $\sim$400 \\
\bottomrule
\end{tabular}
\end{table}

The results presented in Table~\ref{tab:rpi_MMSD} show that semantic segmentation and caption generation dominate the overall computation time, while edge extraction introduces only negligible overhead. Memory usage is estimated based on model footprint and typical runtime buffers during inference. Since the pipeline components are executed sequentially, peak memory consumption is determined mainly by the largest model, which in our case is the BLIP captioning module. We do not include inference latency or memory measurements for the Gemma-based captioning model, as it exceeds the available memory capacity of the Raspberry Pi~5 under CPU-only execution. Nevertheless, this limitation does not really restrict the applicability of the framework, since a lighter application-specific captioning model could be employed at the edge to reduce both memory footprint and CPU cost while still generating sufficiently informative scene descriptions. Thus, the reported values provide an upper bound on resource requirements and support the feasibility of deploying the proposed pipeline on low-end microprocessor IoT devices.


The average inference time for image reconstruction at the receiver, executed on a powerful machine equipped with an NVIDIA RTX 5080 GPU, was measured at 14.47\,s with a standard deviation of 0.04\,s. This result supports the claim that the computational burden of the pipeline is offloaded to the server-side, while edge devices, which are equipped with significantly fewer resources, are responsible only for lightweight processing tasks.


\subsection{{Evaluation of the SAMR Framework}}
\label{sec:eval_masking}

\begin{figure*}
\centering
\subfloat[BPP vs PSNR]{\includegraphics[width=0.32\textwidth]{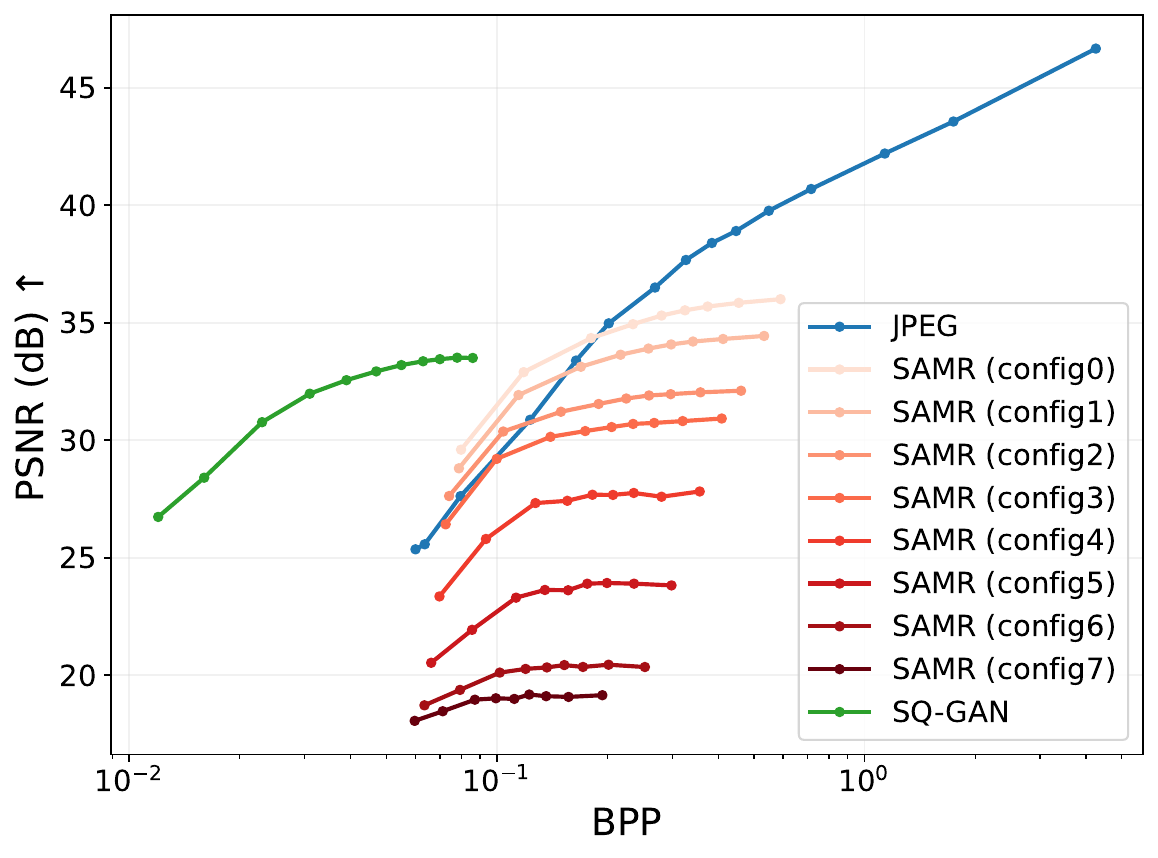}}\hfill
\subfloat[BPP vs MS-SSIM]{\includegraphics[width=0.32\textwidth]{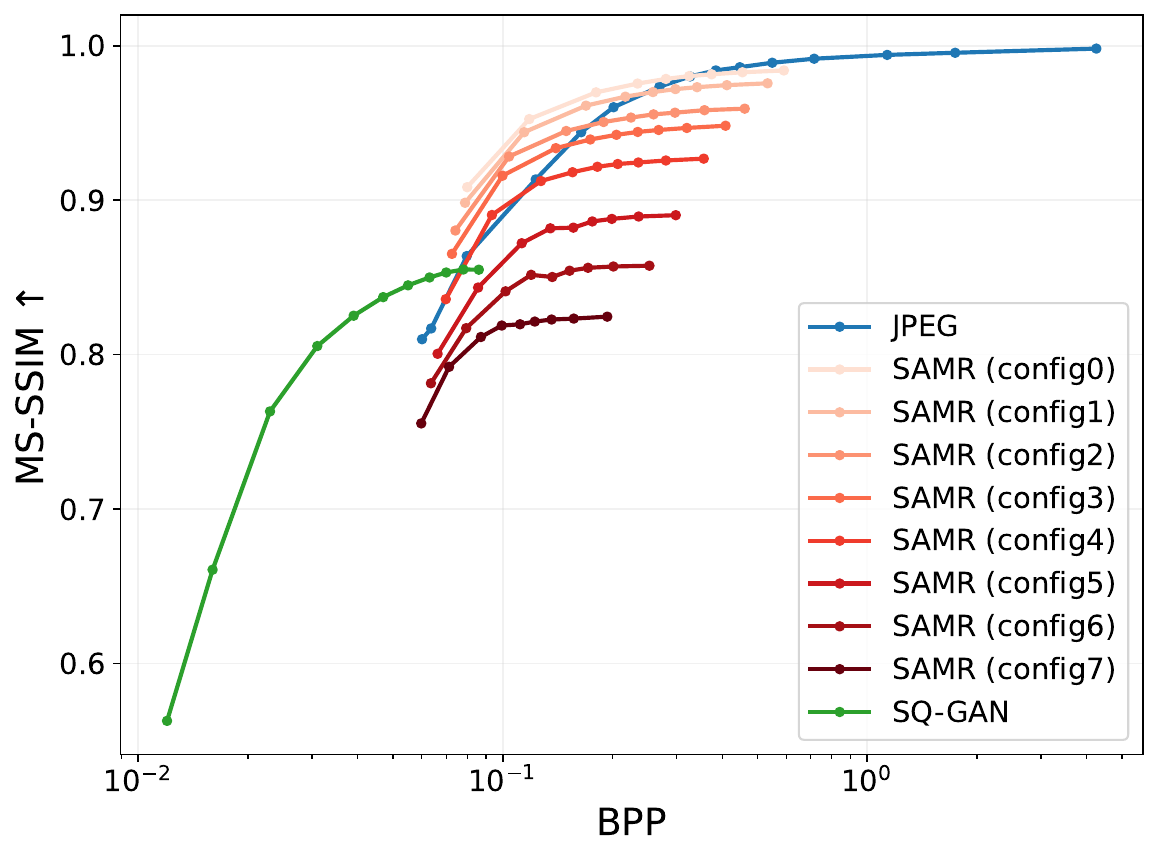}}\hfill
\subfloat[BPP vs LPIPS]{\includegraphics[width=0.32\textwidth]{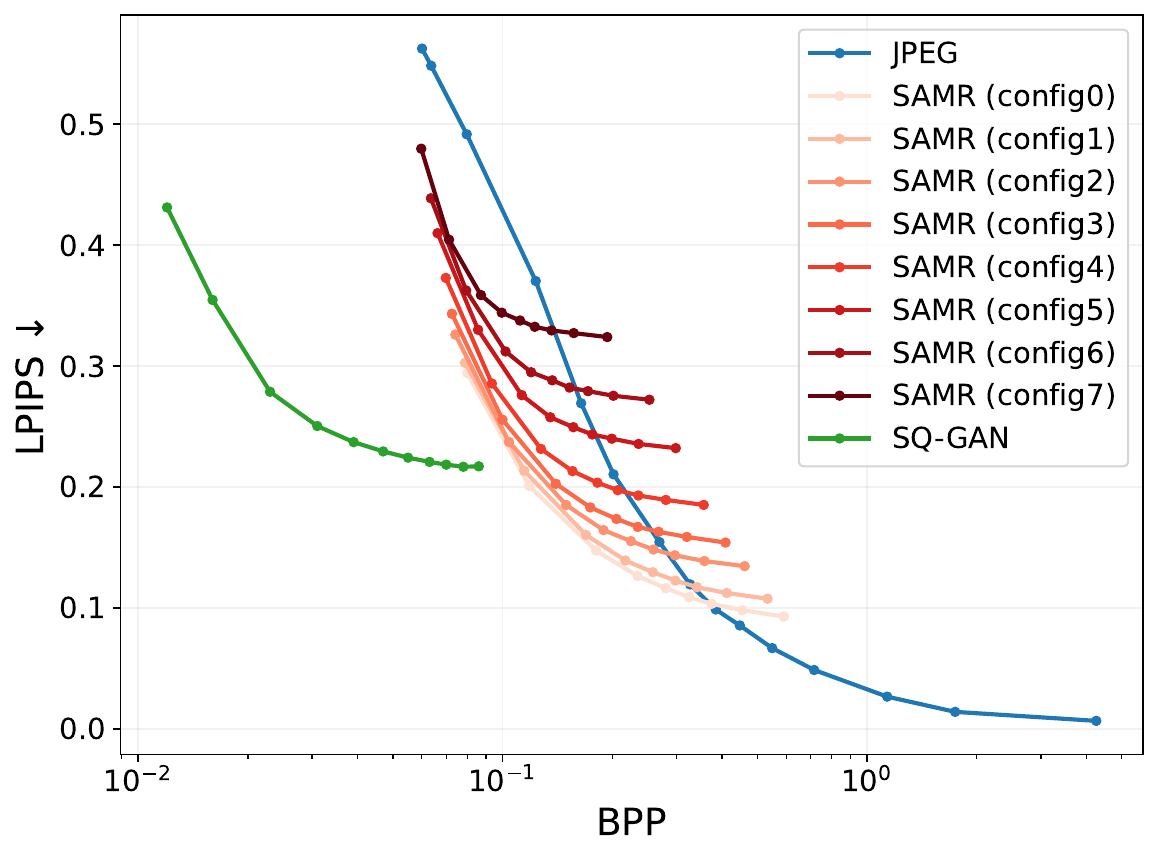}}
\caption{Rate-distortion curves comparing SAMR against standard JPEG and SQ-GAN across three perceptual metrics. The x-axis (BPP) uses a logarithmic scale. Lower BPP indicates higher compression.}
\label{fig:rd-curves}
\end{figure*}

This subsection evaluates the semantic-aware masking approach in terms of quantitative reconstruction accuracy, compression efficiency, and semantic preservation. To provide a comprehensive comparison, we evaluate the proposed method against two baselines that also preserve reconstruction accuracy: standard JPEG compression at 15 quality levels ($Q \in \{1, 3, 5, 10, \ldots, 100\}$), and SQ-GAN~\cite{Pezone2025sqgan}, a recently proposed semantic image communication framework that transmits quantized semantic patches and reconstructs images using a GAN-based decoder.

\subsubsection{Metrics}

We evaluate the performance of SAMR and the other approaches using four metrics:
\begin{itemize}
    \item \textit{Bits per pixel (BPP)} measures compression efficiency as the number of bits required to represent one pixel
    \item \textit{Peak Signal-to-Noise Ratio (PSNR)} quantifies pixel-level reconstruction accuracy
    \item \textit{Multi-Scale Structural Similarity (MS-SSIM)}~\cite{wang2003msssim} captures perceptual similarity across spatial scales.
    \item \textit{Learned Perceptual Image Patch Similarity (LPIPS)}~\cite{Zhang2018lpips} measures perceptual distance using deep features, where lower values indicate better perceptual quality
\end{itemize}

\subsubsection{Rate-Distortion Analysis}

\begin{figure*}
\centering
\subfloat[Full image comparison between the original and the reconstructed pictures of the three methods respectively.]{\includegraphics[width=\textwidth]{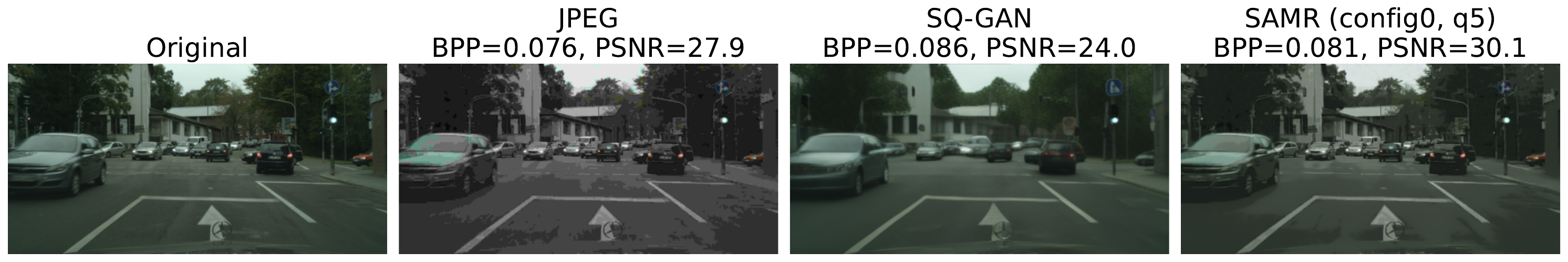}}\\
\subfloat[Zoomed-in center crop ($256 \times 256$ px) comparison between the original and the reconstructed pictures of the three methods respectively.]{\includegraphics[width=\textwidth]{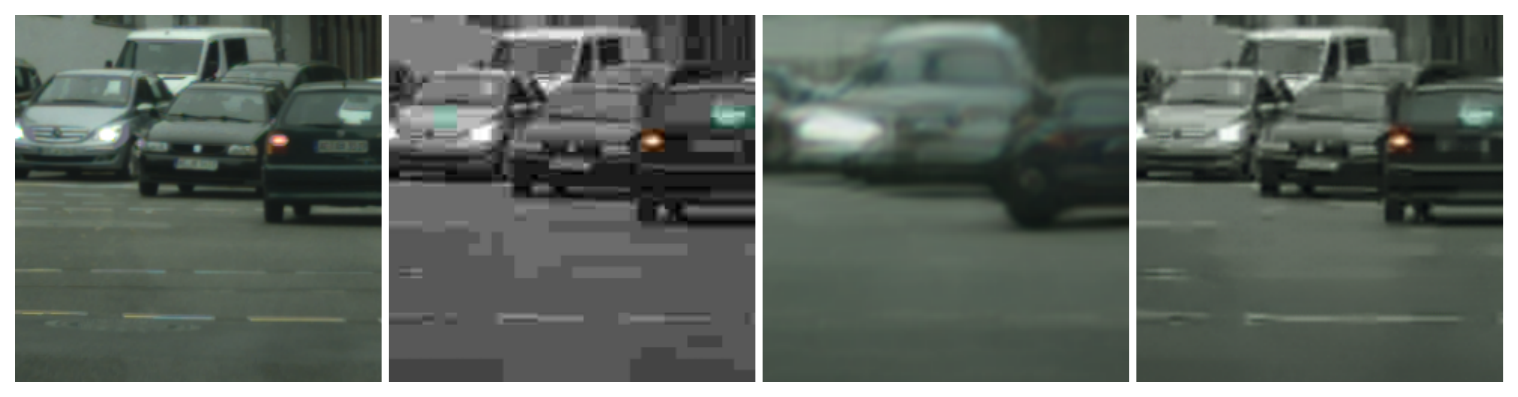}}
\caption{Visual comparison at matched bitrates ($\sim$0.08~BPP). From left to right: original image, JPEG at $Q$=5, SQ-GAN at full masking ($m_x=m_s=1.0$), and SAMR (Config~0, $Q$=5). At comparable bitrates, JPEG exhibits severe blocking artifacts, SQ-GAN produces plausible but low-resolution output, while SAMR preserves structural detail and semantic content at the original resolution.}
\label{fig:visual-full}
\end{figure*}

Figure~\ref{fig:rd-curves} presents the rate-distortion curves for SAMR, standard JPEG, and SQ-GAN across PSNR, MS-SSIM, and LPIPS. The three methods occupy largely non-overlapping bitrate regions. SQ-GAN operates at 0.012--0.086~BPP, transmitting only quantized semantic patches and relying on its GAN decoder for full reconstruction. JPEG spans from roughly 0.06 to over 4~BPP. The proposed method with Config~0 and $Q \in \{5, \ldots, 80\}$ operates between 0.08 and 0.59~BPP, partially overlapping with SQ-GAN only near the lower end at $\sim$0.08~BPP.

\paragraph{Comparison at matched bitrates}
At $\sim$0.08~BPP, JPEG ($Q$=5) achieves PSNR/MS-SSIM/LPIPS of 27.62~dB / 0.864 / 0.492, with severe blocking artifacts. SQ-GAN (full masking, 0.086~BPP) scores 33.50~dB / 0.855 / 0.217, while SAMR (Config~0, $Q$=5, 0.080~BPP) achieves 29.60~dB / 0.908 / 0.294. SQ-GAN's higher PSNR reflects the known tendency of GAN-based smooth reconstructions to minimize pixel-level error while losing fine structural detail---a limitation captured by its lower MS-SSIM (0.855 vs.\ 0.908). Operating at native resolution, SAMR preserves edges, texture, and spatial structure more faithfully, as confirmed by the visual comparison illustrated in Figure~\ref{fig:visual-full}.

\paragraph{Effect of masking configuration and JPEG quality}
The masking configuration is the dominant factor in the bitrate--quality trade-off. From Config~0 to Config~7 at $Q$=5, the bitrate drops only modestly (0.080 to 0.060~BPP, $-$25\%), while PSNR collapses from 29.60 to 18.05~dB and MS-SSIM from 0.908 to 0.755, a highly asymmetric degradation caused by masking semantically important content that the inpainting model cannot recover. Configs 0--2 maintain a controlled trade-off while Configs 4--7 show rapid quality loss. In terms of transferred file size, SAMR achieves a 20\,KB masked file size with Config~4, while the size is 140\,KB with Config~0.

Within a fixed configuration, increasing $Q$ improves quality but with sharply diminishing returns: Config~0 from $Q$=5 to $Q$=80 gains only 6.4~dB in PSNR while multiplying bitrate by $7.4\times$. The largest gain per bit occurs between $Q$=5 and $Q$=10 (+3.3~dB, +0.044 MS-SSIM, at a cost of only 0.038~BPP), making Config~0 / $Q$=10 (0.118~BPP, 32.90~dB, MS-SSIM 0.952) a particularly efficient operating point. Lower $Q$ is generally preferable: the inpainting model compensates for masked-region quality loss, while unnecessarily high $Q$ inflates bitrate without commensurate benefit.

\paragraph{Limitations}
The primary trade-off of SAMR is its minimum achievable bitrate: it cannot match SQ-GAN below $\sim$0.06~BPP, as it must transmit at least the JPEG-compressed unmasked content. Therefore, our method is best suited to scenarios where high quality preservation is the number one target, while image compression is the secondary objective.

\subsubsection{VLM-Based Semantic Preservation Evaluation}

\begin{table}[t]
\centering
\caption{VLM-based (Gemini 27B) semantic preservation evaluation with Config 0 masking.}
\label{tab:llm_judge_masked}
\setlength\tabcolsep{4pt}
\small
\begin{tabular}{l c c c}
\toprule
\textbf{Metric} & \textbf{Overall} \\
\midrule
Position preservation (Q1) & 100.00 \\
Number preservation (Q2)   & 99.80 \\
Joint success rate (Q1 \& Q2) & 99.80 \\
\bottomrule
\end{tabular}
\end{table}

\begin{table}[t]
\centering
\caption{VLM-based perceptual quality comparison at matched bitrate $\sim$0.08~BPP.}
\label{tab:perceptual-comparison}
\small
\setlength\tabcolsep{4pt}
\begin{tabular}{lcc}
\toprule
\textbf{Method} & \textbf{Wins} & \textbf{Win Rate} \\
\midrule
SAMR (Config 0, Q5) & 325 & 65.0\% \\
SQ-GAN ($m_x = m_s = 1.0$) & 175 & 35.0\% \\
JPEG (Q5) & 0 & 0.0\% \\
\bottomrule
\end{tabular}
\end{table}

To evaluate high-level semantic consistency beyond traditional pixel-based metrics, we employ a VLM (Gemini 27B) as an automated judge, as we did with MMSD. For 500 image pairs (original vs. reconstructed), the model assigns a binary score indicating whether key semantic elements, such as vehicles, pedestrians, and overall scene structure, are preserved. As shown in Table~\ref{tab:llm_judge_masked}, SAMR achieves near-perfect semantic retention, with position preservation (Q1) reaching 100\% and number preservation (Q2) 99.80\%. The joint success rate (Q1 \& Q2) remains at 99.80\%, confirming that critical structural and quantitative scene information is consistently maintained despite the high data reduction. 

In addition, we compare SAMR against the standard JPEG and SQ-GAN at matched bitrates ($\sim$0.08~BPP). Using the same VLM for evaluation, the model selects which image better preserves objects, scene structure and overall perceptual quality. As reported in Table~\ref{tab:perceptual-comparison}, the proposed SAMR method achieves 325 wins (65.0\%), compared to 175 wins (35.0\%) for SQ-GAN. These results demonstrate that, at equivalent bandwidth constraints, the semantic reconstruction framework not only preserves high-level meaning but also delivers perceptually preferred outcomes in the majority of cases.

\subsubsection{Edge Deployment and Inference Latency}

\begin{table}[t]
\centering
\caption{Inference Time and Memory Usage on a Raspberry Pi~5.}
\label{tab:rpi_masked}
\setlength\tabcolsep{4pt}
\small
\begin{tabular}{l c c p{1.5cm} c}
\toprule
\multicolumn{5}{c}{\textbf{SAMR}} \\
\multirow{2}{*}{\textbf{Metric}} & \textbf{Segme-} & \multirow{2}{*}{\textbf{Masking}} & \textbf{JPEG} & \multirow{2}{*}{\textbf{Total}} \\
 & \textbf{ntation}& & \textbf{Encoding} & \\
\midrule
Time (s)      & 8.35 & 0.08 & 0.16 & 8.59 \\
Memory (MB)   & $\sim$400 & $\sim$23 & $\sim$3 & $\sim$426 \\
\midrule
\multicolumn{5}{c}{\textbf{SQ-GAN}} \\
\textbf{Metric} & \multicolumn{2}{c}{\textbf{Segmentation}} & \textbf{Encoding} & \textbf{Total} \\
Time (s)      & \multicolumn{2}{c}{7.55} & 13.79 & 21.34 \\
Memory (MB)   & \multicolumn{2}{c}{$\sim$470} & $\sim$438 & $\sim$470 \\
\bottomrule
\end{tabular}
\end{table}

Similarly to the previous approach, to evaluate the deployment feasibility of SAMR on edge devices, we measured its inference latency for each pipeline component on a Raspberry Pi~5 using CPU-only inference.
Table~\ref{tab:rpi_masked} presents the inference time and memory usage of the SAMR and SQ-GAN. For SAMR, the segmentation stage dominates both computation time ($\sim$8\,s) and memory consumption ($\sim$400\,MB), while masking and JPEG encoding add only minimal overhead. Overall, the complete pipeline executes in 8.59\,s with a max memory footprint of about 400\,MB, largely driven by the segmentation model. On the other hand, SQ-GAN needed approximately $\sim$14 additional seconds for the encoding part resulting in a much higher total computation time. 

The inference time of SAMR at the receiver running on the RTX 5080 GPU was measured at approximately 20\,ms with a deviation of 0.4\,ms. The same experiment on a 10-core Intel i9 CPU took approximately 1.8\,s and 3.2\,GB of RAM per image reconstruction. SQ-GAN's reconstruction was roughly 3 times slower.

\section{Conclusions \& Future Work}
\label{sec:conclusion}

In this paper, we presented two complementary semantic communication frameworks, MMSD and SAMR, for traffic monitoring at the edge. MMSD enables high compression and scene confidentiality by transmitting compact semantic representations, including segmentation maps, edge maps, and textual descriptions, while SAMR focuses on preserving high compression ratio together with higher visual fidelity through semantic-aware masking and generative reconstruction at the cost of weaker confidentiality. The two approaches follow an asymmetric edge-server design and can be flexibly combined depending on the application requirements. Experimental results demonstrate that both approaches achieve $\ge$99\% data reduction while maintaining strong semantic consistency. SAMR provides a better quality/compression trade-off and supports on-demand higher-fidelity reconstruction with minimal additional edge-side overhead.

Future work will explore several directions. First, we plan to investigate adaptive encoding strategies for edge maps, where the amount of structural information (e.g., edge density or thresholds) can be dynamically tuned to further balance compression and reconstruction quality. Second, lightweight and application-specific vision-language models can be designed to reduce the computational cost of caption generation on resource-constrained devices. Finally, we will extend the proposed frameworks to other application domains beyond traffic monitoring, such as aerial imaging and smart city sensing, where similar trade-offs between bandwidth, semantics, and visual fidelity are required.

\bibliographystyle{unsrt}
\bibliography{biblist}

\end{document}